\documentclass[10pt,twocolumn,letterpaper]{article}

\usepackage[pagenumbers]{cvpr} %

\usepackage[dvipsnames]{xcolor}

\definecolor{cvprblue}{rgb}{0.21,0.49,0.74}
\usepackage[pagebackref,breaklinks,colorlinks,citecolor=cvprblue]{hyperref}

\usepackage{amssymb}%
\usepackage{pifont}%
\usepackage{booktabs}
\usepackage{graphicx}
\usepackage{booktabs}
\usepackage{multirow}
\usepackage{array}
\newcolumntype{P}[1]{>{\centering\arraybackslash}p{#1}}
\usepackage{listings}
\usepackage{svg}
\usepackage{scalerel}
\usepackage{oplotsymbl}
\usepackage{wasysym}
\usepackage{tikz}
\usetikzlibrary{shapes}
\usetikzlibrary{shapes.geometric}
\usepackage{scalerel}[2016/12/29]
\usepackage{xcolor,colortbl}
\usepackage{titletoc}

\newcommand{\triup}{%
\begin{tikzpicture}%
\draw [line width=0.3ex] (0,0) -- (0,1ex) {};%
\draw [line width=0.3ex] (0,0) -- (-0.8ex,-0.5ex) {};%
\draw [line width=0.3ex] (0,0) -- (0.8ex,-0.5ex) {};%
\end{tikzpicture}%
}

\newcommand{\tridown}{%
\begin{tikzpicture}%
\draw [line width=0.3ex]  (0,0) -- (0,-1ex) {};%
\draw [line width=0.3ex]  (0,0) -- (0.8ex,0.5ex) {};%
\draw [line width=0.3ex] (0,0) -- (-0.8ex,0.5ex) {};%
\end{tikzpicture}%
}

\newcommand{\btimes}{%
\begin{tikzpicture}%
\draw [line width=0.3ex]  (0,0) -- (1ex,1ex) {};%
\draw [line width=0.3ex]  (1ex,0) -- (0,1ex) {};%
\end{tikzpicture}%
}

\newcommand{\mdblksquare}{%
\begin{tikzpicture}%
\fill (-.5ex,-.5ex) rectangle (.5ex,.5ex) {};%
\end{tikzpicture}%
}

\newcommand{\blackoctagon}{%
    \tikz{
        \node[fill,scale=.75,regular polygon, regular polygon sides=8](){};
    }
}
\newcommand{\varhexagonblack}{\tikz{\node[fill,scale=.75,regular polygon, regular polygon sides=6](){};}}
\newcommand{\smallblacktriangleleft}{\tikz{\node[fill,scale=0.5,isosceles triangle,isosceles triangle apex angle=60,rotate=180](){};}}
\newcommand{\smallblacktriangleright}{\tikz{\node[fill,scale=0.5,isosceles triangle,isosceles triangle apex angle=60,rotate=0](){};}}

\def\confName{CVPR}
\def\confYear{2024}

\title{Can Biases in ImageNet Models Explain Generalization?}
\author{Paul Gavrikov \\
IMLA, Offenburg University\\
{\tt\small paul.gavrikov@hs-offenburg.de}
\and Janis Keuper\\
IMLA, Offenburg University and University of Mannheim\\
{\tt\small keuper@imla.ai}
}

\definecolor{c_stylized}{RGB}{31,119,180}
\definecolor{c_adversarial_training}{RGB}{255,127,14}
\definecolor{c_training_recipes}{RGB}{44,160,44}
\definecolor{c_freezing}{RGB}{214,39,40}
\definecolor{c_augmentation}{RGB}{148,103,189}
\definecolor{c_contrastive}{RGB}{140,86,75}
\definecolor{c_baseline}{RGB}{0,0,0}

\begin{document}
\maketitle
\begin{abstract}
The robust generalization of models to rare, in-distribution (ID) samples drawn from the long tail of the training distribution and to out-of-training-distribution (OOD) samples is one of the major challenges of current deep learning methods.    
For image classification, this manifests in the existence of adversarial attacks, the performance drops on distorted images, and a lack of generalization to concepts such as \textit{sketches}. The current understanding of generalization in neural networks is very limited, but some biases that differentiate models from human vision have been identified and might be causing these limitations. Consequently, several attempts with varying success have been made to reduce these biases during training to improve generalization. We take a step back and sanity-check these attempts. Fixing the architecture to the well-established ResNet-50, we perform a large-scale study on 48 ImageNet models obtained via different training methods to understand how and if these biases - including shape bias, spectral biases, and critical bands - interact with generalization.
Our extensive study results reveal that contrary to previous findings, these biases are insufficient to accurately predict the generalization of a model holistically.\\
We provide access to all checkpoints and evaluation code at \url{https://github.com/paulgavrikov/biases_vs_generalization/}
\end{abstract}

\DeclareRobustCommand{\emojirobot}{%
  \begingroup\normalfont
  \includegraphics[height=\fontcharht\font`\B]{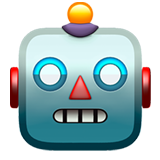}%
  \endgroup
}

\DeclareRobustCommand{\emojihuman}{%
  \begingroup\normalfont
  \includegraphics[height=\fontcharht\font`\B]{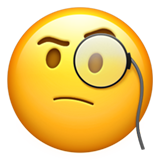}%
  \endgroup
}

\newcommand{\cbar}{IN-\={C}}
\begin{figure}
    \centering
    \includegraphics[width=\linewidth]{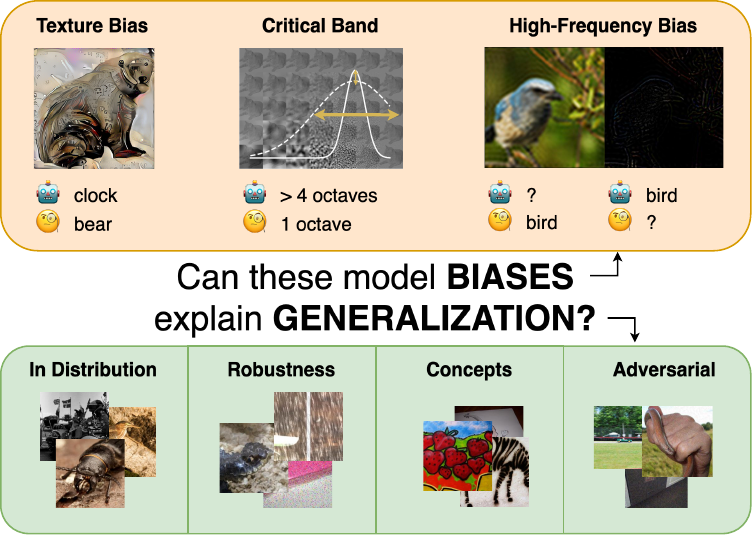}
    \caption{\textbf{We study the influence of three selected biases that separate models \emojirobot{} from humans \emojihuman{} on the generalization of ImageNet models. Our study suggests that no single bias correlates with generalization in a holistic sense.} We measure the texture/shape bias \cite{geirhos2018imagenettrained}, critical band \cite{subramanian2023spatialfrequency}, and low/high-frequency spectral biases \cite{Wang_2020_CVPR} on 48 models and correlate these biases against generalization that we measure on several benchmarks belonging to four categories: in distribution, robustness, conceptual changes, and adversarial robustness.}
    \label{fig:teaser}
\end{figure}
\section{Introduction}
\label{sec:introduction}
Artificial neural networks have achieved outstanding performance in various tasks, particularly excelling in vision tasks like image classification - commonly benchmarked on ImageNet \cite{imagenet}. While accuracy on ImageNet has dramatically improved in the last 15 years \cite{alexnet,resnet,dosovitskiy2021an}, with models nearly matching human performance, a critical issue remains: these models often fail to generalize beyond the specific data distribution they were trained on. This fragility becomes evident when faced with distribution shifts, such as changes in weather conditions \cite{NEURIPS2020_d8330f85}, digital artifacts \cite{hendrycks2019commoncorruptions}, or even replacing photos objects with artistic renditions \cite{hendrycks2021rendition}. On an abstract level this can be unified by the vulnerability to adversarial attacks \cite{SzegedyZSBEGF13,biggio2013evasion}, where carefully crafted image perturbations can fool models while remaining imperceptible to humans. This raises serious concerns about deploying such models in safety-critical applications where human lives are at stake \cite{Finlayson2019, cao2019}.

Seeking explanations for this lack of generalization, researchers have identified various biases that differentiate model from human perception. Three recent examples that have received significant attention are the \textit{texture/shape bias} (models rely heavily on texture for object recognition, whereas humans prioritize shape information) \cite{geirhos2018imagenettrained}, \textit{critical band bias} (models utilize wider critical bands than humans for recognizing ImageNet objects) \cite{subramanian2023spatialfrequency}, and \textit{high-frequency bias} (models exhibit a strong dependence on high-frequency information and perform poorly on low-pass filtered data) \cite{Wang_2020_CVPR}.

Following the intuition that aligning these biases with human perception might improve generalization, several studies have explored regularization techniques during training \cite{li2021shapetexture,lopes2020improving,SaikiaSB21,lukasik2023improving}. However, the fundamental question (visualized in \cref{fig:teaser}) remains: \textit{how well do these biases actually correlate with the ability to generalize?} 

In this study, we take a step back and analyze these biases in ImageNet models, investigating their connection to generalization performance under various aspects. A cornerstone of this analysis is that we explore changes \textit{in training} (\eg, augmentation, contrastive learning, and others), \textit{while keeping the model architecture constant} to isolate the effect of inductive biases. We assess generalization by measuring performance on different datasets modeling in-distribution data, corrupted data, conceptual changes, and under adversarial attacks.

This investigation aims to shed light on the true impact of these biases and inform the development of more robust and generalizable models.

\noindent\textbf{We summarize our contributions as follows:}
\begin{itemize}
    \item We systematically evaluate a diverse set of 48 models trained using various techniques for ImageNet classification. We assess their performance across multiple generalization benchmarks, including in-distribution and out-of-distribution tests related to robustness, concept generalization, and adversarial robustness. Then, we examine correlations between these benchmarks and biases like shape bias, spectral biases, and characterizations of the critical band.
    \item Our analysis highlights that previous studies linking generalization to tested biases often examined only partial aspects (\eg, specific training or benchmarks), introduced confounders by inductive biases, or encountered challenges due to noisy measurements originating from system noise and incompatible data transformation pipelines.
    \item \textbf{Our findings reveal several key insights}: \textbf{i)} none of the tested biases can singularly explain generalization - sometimes they are even negatively correlated with human perception; \textbf{ii)} most of the previously observed correlations show many outliers; \textbf{iii)} in our evaluation, some biases exhibit no correlation at all; \textbf{iv)} we find a surprising correlation between a high-frequency bias and generalization.
\end{itemize}

\section{Related Work}
\label{sec:related_work}

\paragraph{Texture/Shape Bias.}
\citet{geirhos2018imagenettrained} discovered that CNNs primarily rely on texture cues to identify objects (in ImageNet). This is in stark contrast to the strong shape-biased decision observed in human perception. They argue that improving the shape bias may be linked to improved robustness and show that (pre-)training on a dataset without discriminative texture information indeed improves robustness on some benchmarks including common corruptions \cite{hendrycks2019commoncorruptions}. Following this observation, multiple works have attempted to understand origins of this bias \cite{hermann2020origins,islam2021shape}, improve shape bias in training \cite{li2021shapetexture,pmlr-v119-shi20e,lukasik2023improving}, or study representations beyond ImageNet-CNNs \cite{NEURIPS2021_c404a5ad,jaini2023intriguing,gavrikov2024vision,dehghani23a,Geirhos2021_modelsvshumasn}. Prior works have also shown a correlation between adversarial-training (AT) \cite{madry2018towards} and shape bias \cite{Geirhos2021_modelsvshumasn,Gavrikov_2023_CVPRW}.
However, an improved shape bias seems not to be the cure-all for generalization: \cite{mummadi2021does} designed an augmentation technique that improves shape bias but not robustness to common corruptions \cite{hendrycks2019commoncorruptions} and, thus, shows a counter-example to reject a causal connection at least for this aspect of generalization. In our study, we test the shape bias beyond common corruptions to derive a more holistic understanding of this bias in the context of generalization.

\paragraph{Spectral Bias.}
Natural signals, such as images \cite{Ruderman1994TheSO} and sounds \cite{Singh2004}, concentrate most information in low-frequency bands. As such, it seems to be reasonable to draw predictive cues from this band. Contrary, \citet{Wang_2020_CVPR} demonstrated that image classification models reach high accuracies on high-pass filtered images that mostly resemble noise and should not carry discriminative information. Low-pass filtered images easily recognizable by humans, on the other hand, are poorly recognized by the same models. While this seems to be more extreme on low-resolution images, higher-resolution models such as ImageNet classifiers also unreasonably heavily rely on cues from high-frequency (HF) bands \cite{AbelloHW21}. \cite{Wang_2020_CVPR} also link this \textit{spectral} bias to adversarial robustness \cite{SzegedyZSBEGF13,biggio2013evasion} and show that adversarial training (AT) \cite{madry2018towards} results in a reversed bias. Indeed, perturbations generated by adversarial attacks on ImageNet seem to particularly target HF bands \cite{dong2019fourier}. A similar connection to non-adversarial robustness can be made, as most common corruptions \cite{hendrycks2019commoncorruptions} primarily target HF bands \cite{dong2019fourier}.
This bias has received attention in the robustness community, where some hope to improve robustness by stronger low-frequency biases \cite{lopes2020improving,SaikiaSB21,lukasik2023improving}. We seek to understand how spectral biases, in particular, low/high-frequency biases affect generalization.

\paragraph{Critical Band.}
Recently, \citet{subramanian2023spatialfrequency} have found another bias that separates human from model vision in the \textit{critical band}. This band defines what spatial-frequency channels are used to detect objects. On ImageNet, humans use a one-octave wide channel, yet models measure significantly wider channels, making them more sensitive to noise perturbations in spectral bands that do not affect human vision. The study showed a strong correlation between the parameters of the critical band and shape bias, as well as the adversarial robustness of adversarially-trained networks. Surprisingly, adversarial training (AT) seems to further increase the bandwidth and, thus, increases misalignment. We propose a few optimizations to the underlying test process and as for the previous biases, aim to understand possible connections to generalization beyond adversarial robustness/training.

\section{Methodology}
\label{sec:method}

In our study, we measure models' ability to generalize and correlate this to the biases that separate human and model vision. The following paragraphs show details about how we measure biases, generalization, and correlation, as well as our tested models.

\subsection{Measuring Biases}

 \paragraph{Texture/Shape Bias.} 
 We measure the texture/shape bias on the \textit{cue-conflict} dataset \cite{geirhos2018imagenettrained}. This dataset consists of ImageNet samples where shape and texture cues belong to different (\textit{conflicting}) ImageNet super-classes (\eg, a bear with clock skin as shown in \cref{fig:teaser}). We report the \textit{shape bias} which is defined by the ratio of shape to all correct decisions on the dataset - with a value of 0/1 indicating a texture/shape bias, respectively. Humans achieve a strong average shape bias of $0.96$ but most models are severely texture biased.

\paragraph{Spectral Bias.} 
To understand the spectral bias we measure the performance of bandpass-filtered ImageNet samples, akin to \cite{Wang_2020_CVPR}. We differentiate between low and high-frequency biases by filtering the test data at different cutoff frequencies. For instance, a low-pass filter with a cutoff of 30\% will contain the lowest 30\% of the Fourier spectrum. To compensate for different test performances of the models on the clean data, we report the \textit{ratio} of the accuracy on the filtered data to the clean accuracy. In some cases, this may result in ratios $> 1$, \ie, the accuracy on filtered data may exceed the original validation accuracy. As low/high-frequency bands are ill-defined we average the performance observed at 10, 20, 30, and 40\% cutoffs along the low/high-pass filter tests to report the \textit{low/high-frequency bias}, respectively.

\paragraph{Critical Band.} 

Following the methodology of \citet{subramanian2023spatialfrequency}, we measure the critical band by insertions of Gaussian noise to contrast-reduced and gray-scaled ImageNet samples with varying strength at different frequencies. Areas where the accuracy falls below a predefined threshold mark the critical band. The parameters of the band are obtained by the fitting of a Gaussian characterized by the \textit{bandwidth (BW)}, \textit{center frequency (CF)}, and \textit{peak-noise sensitivity (PNS)}. A caveat of the original test is, that the critical band is determined by conditions where the accuracy (on a 16-class ImageNet subset) drops below 50\%. However, adversarially-trained models perform poorly on contrast-reduced images \cite{Geirhos2021_modelsvshumasn,Gavrikov_2023_CVPRW}. To compensate for this fact, we \textit{normalize} the accuracy by the performance on the contrast-reduced images before applying noise. Furthermore, the original test only uses a few ImageNet samples for each condition. Instead, we use all ImageNet validation samples. Finally, we determine the accuracy against all 1,000 classes as opposed to 16 super-classes. Taken together this results in a more computationally expensive but also more accurate measurement. In \cref{app_sec:criticalband}, we also show results with the original test routine - with and without normalization.

\subsection{Measuring Generalization}
To benchmark generalization we loosely follow the suggestions of prior work \cite{hendrycks2021rendition} and rely on multiple semantically-different benchmarks to faithfully assess facets. Our benchmarks can be separated into \textbf{four categories}: \textit{in-distribution (ID)} datasets with considerable statistical similarity to the ImageNet training set, \textit{robustness} datasets that apply various corruptions to ImageNet samples, datasets challenging models to recognize objects based on \textit{concepts} (\eg sketches), and \textit{adversarial attacks} to test robustness adaptively. For all benchmarks we evaluate the top-1 accuracy using a resolution of $224^2$ px and \textit{consistent} transformation pipelines on the \textit{same system} to avoid the risk of falsifications due to system noise \cite{wang2021systematicnoise}. We report the average over the individual benchmarks for each of the four categories.
In the following, we briefly introduce the benchmarks but refer the reader to \cref{app_sec:dataset_details} for full details.

\paragraph{In Distribution (ID).} We use the \textit{ImageNet (IN)} \cite{imagenet} validation set from the ILSVRC2012 challenge, \textit{ImageNet v2 (IN-v2)} \cite{recht19imagenet2} which is a newer test set sampled a decade later following the methodology of the original curation routine, and \textit{ImageNet-ReaL (IN-ReaL)} \cite{beyer2020imagenet} which is a re-annotated version of the original ImageNet validation set. IN-ReaL assigns multiple labels per image and contains multiple corrections of the original annotations.

\paragraph{Robustness.} We benchmark robustness on \textit{ImageNet-C (IN-C)} \cite{hendrycks2019commoncorruptions} which applies 19 synthetic corruptions under five levels of severity and its successor \textit{ImageNet-\={C} (\cbar)} \cite{mintun2021cbar} which extends this set by 10 additional corruptions which were chosen to be perceptually dissimilar but conceptually similar to the corruptions in IN-C. Additionally, we use \textit{ImageNet-A (IN-A)} \cite{hendrycks2021nae} which contains samples from the full ImageNet that are naturally hard to classify and, thus, pose natural adversarial examples.

\paragraph{Concepts.} To understand model capabilities to recognize objects under conceptual changes, we use \textit{ImageNet-Renditions (IN-R)} \cite{hendrycks2021rendition} representing ImageNet objects through cartoons, paintings, toys, etc., \textit{ImageNet-Sketch (IN-S)} \cite{wang2019learning} consisting of hand-drawn sketches of ImageNet objects, and \textit{Stylized ImageNet (SIN)} \cite{geirhos2018imagenettrained} containing ImageNet validation images that have been stylized using different artistic filters to destroy the correlation between texture information and class labels.

\paragraph{Adversarial Robustness.} We apply a Project Gradient Descent (PGD) \cite{madry2018towards} attack on the ImageNet validation set to adaptively evaluate robustness. Contrary to the previous benchmarks, this benchmark is \textit{adaptive} as it can continuously generate weaknesses for a given model. In theory, a model might overfit a static dataset due to its limited number of test samples; however, an (ideal) adaptive benchmark remains unaffected by this limitation.\\

\subsection{Models}

We conduct our main analysis on models based on the ResNet-50 architecture \cite{resnet} that were trained on ImageNet \cite{imagenet}\footnote{\cref{app_sec:vit} also contains results for ViT-B/16 models.}. Fixing the architecture is crucial for our study, as each architecture has its own inductive biases that may influence results. Thus, to disentangle the inductive bias from the influence on generalization we ensure that all models use an identical architecture. This also means avoiding even minor changes like different activation functions as they can drastically affect the performance and introduce confounders to our analysis \cite{xie2021smooth}. Limiting the study to ResNet-50 has additional benefits: the architecture is highly popular and comes without bells and whistles. This status also ensures that a variety of techniques have been evaluated for this architecture, specifically. The architecture also represents architectures used in real-life applications, where training from scratch is necessary but computational resources are limited. Further, we ensured no model was trained with external data to avoid target leakage on our benchmarks. Most models were also not specifically optimized for the tested biases which provides an opportunity to understand if and to what extent (aligned) biases arise through changes in training.

To represent changes in generalization, we use 48 models that differ \textit{in training}. We loosely sort these models into seven categories that we color-code. Throughout the study, we use different marker symbols to differentiate individual models. Marker sizes of adversarially-trained models correlate with the attack budget $\epsilon$ during training. A full model legend including all measurements can be found in \cref{app_sec:model_tables}.

Our tested models belong to the following categories:\\
\textcolor{black}{$\bullet$} a \textit{baseline} model following the  original training \cite{resnet};\\ 
\textcolor{c_augmentation}{$\bullet$} \textit{augmentation} techniques \cite{hendrycks2020augmix,hendrycks2022pixmix,hendrycks2021rendition,jaini2023intriguing,Muller_2023_ICCV,modos2022prime,li2021shapetexture,erichson2022noisymix};\\
\textcolor{c_stylized}{$\bullet$} \textit{(pre)training} on \textit{SIN} which can be seen as an extreme form of augmentation that breaks correlations between textures and class labels (\textsc{ShapeNet}) \cite{Geirhos2021_modelsvshumasn};\\
\textcolor{c_adversarial_training}{$\bullet$} \textit{adversarial training} \cite{madry2018towards} against a PGD adversary with increasing attack budget $\epsilon$ under $\ell_2$ and $\ell_\infty$-norm \cite{salman2020transfer}. Technically, this too could be seen as augmentation, but we will show significant differences for this group of models;\\
\textcolor{c_contrastive}{$\bullet$} \textit{contrastive} learners \cite{Chen_2021_ICCV,caron2021emerging,caron2020swav,chen2020simclrv2} with supervised finetuning of the classification-head;\\ 
\textcolor{c_training_recipes}{$\bullet$} improved \textit{training recipes} in \textsc{timm} \cite{rw2019timm} based on the findings of \citet{wightman2021resnet} and \textsc{PyTorch} \cite{PyTorch,BibEntry2023Nov};\\
\textcolor{c_freezing}{$\bullet$} a model with \textit{randomly weighted} convolution filters \cite{gavrikov2023power}

\subsection{Measuring correlations}
We measure correlations via the Spearman rank-order correlation coefficient $r$. Due to the low sample size ($N \ll 500$), we also assess the statistical significance by obtaining a $p$-value from a two-sided permutation test. $p$ is given by the ratio of generated $r$ larger than the baseline $r$ on all samples. We consider results as significant if $p<0.05$. 

\begin{figure*}
    \centering
    \includegraphics[width=\linewidth]{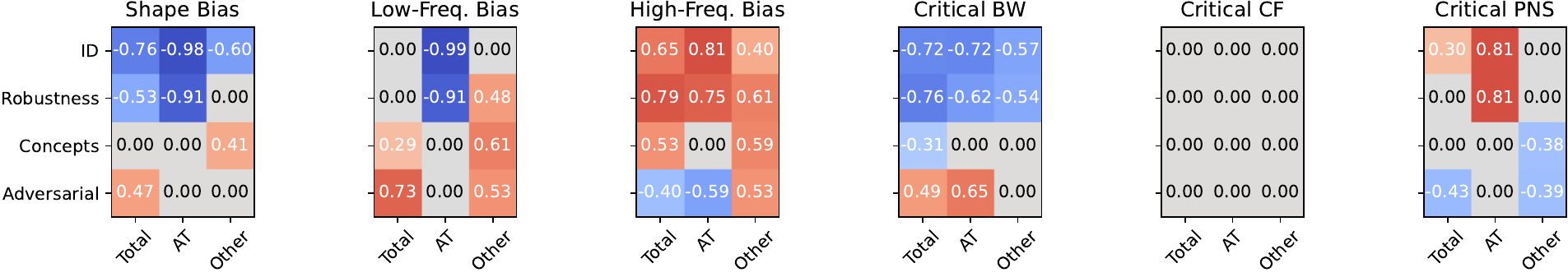}
    \caption{\textbf{Biases often only correlate with specific aspects of generalization or model groups.} We measure Spearman $r$ correlations on all models (\textit{Total}) and separately on adversarially-trained (\textit{AT}), and all \textit{other} models, as there is often a different trend. Non-significant correlations with $p\geq 0.05$ are set to 0. Please note that $r$ does not capture non-monotonic relations.}
    \label{fig:heatmaps_corr_sbs}
\end{figure*}
\section{Limitations of previous studies}
\label{sec:prev_results}

Previous studies have already correlated these biases with (some aspects of) generalization \cite{Wang_2020_CVPR,subramanian2023spatialfrequency,Geirhos2021_modelsvshumasn,geirhos2018imagenettrained}. In this section, we discuss limitations in these studies and show why their findings may not transfer to a more holistic view of generalization.

\paragraph{Inductive biases of architectures were not ablated.}
The architecture plays a significant role in robustness \cite{Tang2021} and, thus, can also be expected to affect a more holistic view of generalization. As such, previously discovered correlations may be spurious concerning the bias itself and just an effect of the architecture. To suppress this potential confounder our model zoo is limited to ResNet-50 models without any modifications to the original architecture as defined in \cite{resnet}. As a matter of fact, we observe that many correlations no longer hold when we fix the architecture.
\begin{figure}
    \centering
    \includegraphics[width=0.9\linewidth]{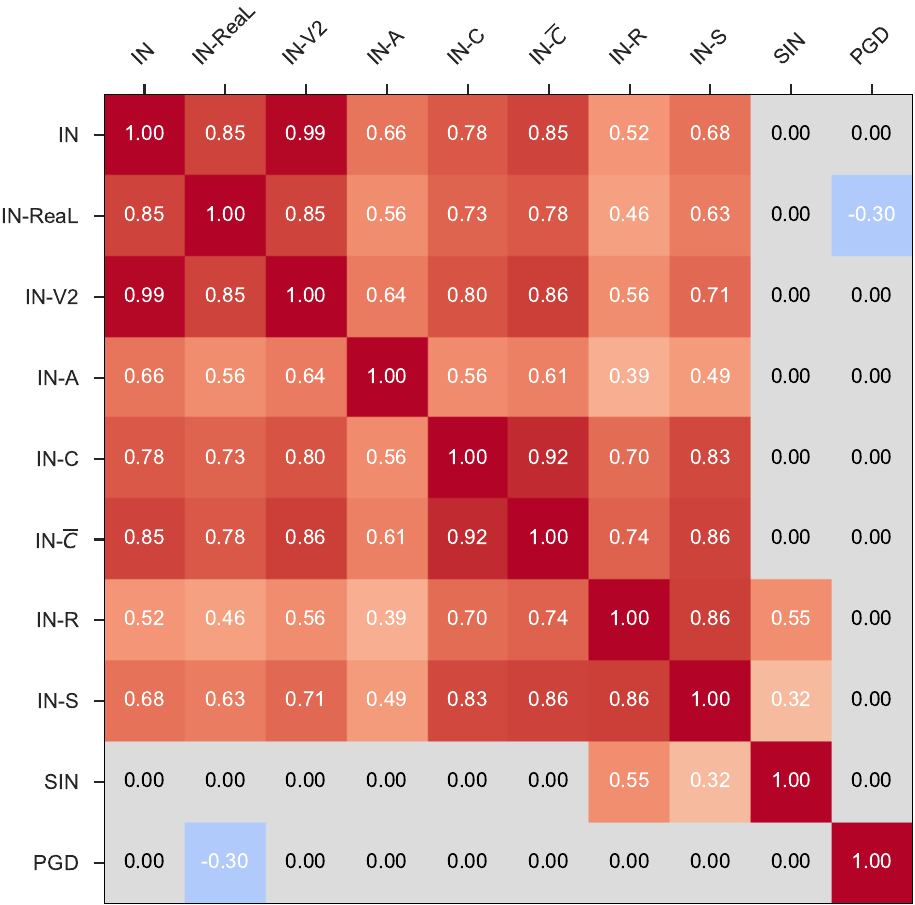}
    \caption{\textbf{Many benchmarks show notable positive correlations between each other - except SIN and PGD (adversarial attack).} Correlations measured by Spearman $r$. We set non-significant correlations with $p\geq 0.05$ to 0.}
    \label{fig:heatmap_benchmarks_no_sslswsl}
\end{figure}
\paragraph{Findings on subsets do not generalize.}
It is well known that correlation does not imply causation and, thus, it may not be surprising that we find that many observed correlations no longer hold for our study. Common causes include (i) benchmarking only a specific type of training, \eg, adversarial training. Yet, correlations for adversarial training may be differently strong or even reversed compared to other forms of training (for example see \cref{fig:heatmaps_corr_sbs}); and/or (ii) using only one benchmark, \eg, adversarial attacks. Performance on many ImageNet-benchmarks is moderate to strongly correlated \cite{NEURIPS2020_d8330f85}, which we also confirm in our study (\cref{fig:heatmap_benchmarks_no_sslswsl}) but clear outliers exist. For instance, adversarial robustness (here: PGD) is not correlated to most other benchmarks (\cref{fig:heatmap_benchmarks_no_sslswsl}) and generally might be divergent to other forms of robustness \cite{liu2023comprehensive,tsipras2018robustness,raghunathan019adversarial,stutz2019}. Our study focuses on generalization beyond but including adversarial robustness and is performed on various models including adversarially-trained ones.

\paragraph{Inconsistent benchmarking.} 
While the surrounding details of benchmarks in previous studies are not always clear, a common procedure is to simply copy benchmark results reported in publications. However, concerningly, we found that some authors produced results using inconsistent transformation pipelines during testing resulting in non-comparable results. 
Additionally, even system noise can lead to significant errors in evaluations \eg, due to low-level implementation details in data loaders such as JPEG decompression or subsampling techniques \cite{wang2021systematicnoise}. These issues advocate for benchmarking under comparable conditions to obtain comparable results. Therefore, we produce all benchmarks on the same system, using the same environment, and identical test transformations for all datasets.

\section{Can these biases explain generalization?}
\label{sec:results}
\begin{figure}
    \centering
    \includegraphics[width=\linewidth]{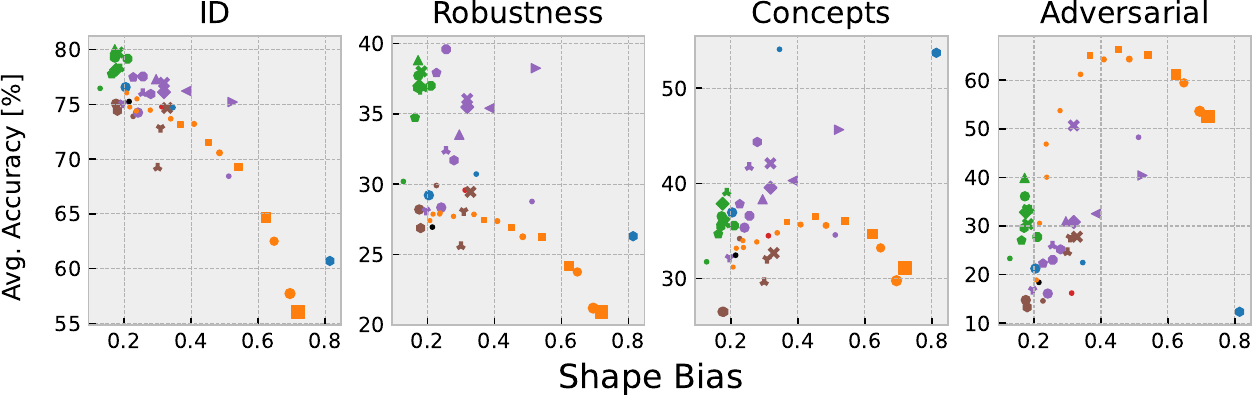}
    \caption{\textbf{Shape Bias vs.~Generalization.} A value of 0 indicates a texture bias, and 1 is a shape bias.}
    \label{fig:scatter_shapebias}
\end{figure}

In this section, we apply the previously introduced methodology on each bias individually, to understand if and to what extent aligning these biases can help with generalization. \Cref{fig:heatmaps_corr_sbs} shows an overview of the \textit{statistical} correlation for all tested biases. Additionally, we inspect scatterplots (\cref{fig:scatter_shapebias,fig:scatter_spatial,fig:scatter_cb}) to detect non-monotonic correlations missed by Spearman $r$ and pay attention to whether outliers exist. 
\subsection{Shape Bias}
\label{subsec:shape_bias}
For the shape bias, we observe a strong \textbf{negative} correlation to ID tests (\cref{fig:scatter_shapebias}) - suggesting that increasing misalignment with human vision increases ID accuracy. While it has been shown before, that ImageNet-CNNs are strongly texture-biased and ImageNet can be well separated by texture alone \cite{geirhos2018imagenettrained}, the monotonic decrease in performance with increasing shape bias without significant outliers is quite surprising. While this does not prove causality, it raises the question of whether strongly shape-biased models can achieve significant performance on ImageNet (without relying on extra data). Similarly, we only note significant improvements in IN-A on strongly texture-biased models (see \cref{app_sec:detailed_plots} for a plot). Usually, poor performance on IN-A is linked to spurious cues \cite{hendrycks2021nae} and one may assume that these are related to textures (similar to adversarial attacks on fragile features \cite{ilyas2019bugsnotfeat}). However, it seems like these cues are more shape-related and a certain degree of texture bias is necessary to overcome them. But neither does a texture bias guarantee the best performance, as the strongest texture-biased model (\textsc{timm (A3)} \cite{wightman2021resnet}) with a shape bias of $0.13$ still performs 10.4\% worse on IN-A than the best model (\textsc{torchvision (v2)}) with a (higher) shape bias of $0.17$. 

On other aspects of generalization, we often obtain unrelated trends between adversarially trained and other models. For adversarially-trained (AT) models, we observe a curvilinear (inverted ``U'') relation (which is not captured by the Spearman $r$) to all aspects of generalization which also seems to correlate with the attack budget during training (pay attention to the marker size). For concepts and adversarial robustness, we observe the best performance by AT models that balance shape and texture bias ($\sim 0.5$).
There are no notable differences in the trends between models trained on $\ell_2$ and $\ell_\infty$-norm attacks. Non-AT models only show a (moderate) correlation to concepts.

Our results extend previous findings showing that shape bias is not causally correlated with IN-C performance \cite{mummadi2021does} to a much broader view of generalization, and contrast the claims about improved robustness in \cite{geirhos2018imagenettrained}. For adversarial robustness, shape bias may define a ceiling where a balanced representation of shapes and textures works best. However, the model set of top performers exclusively consists of adversarially-trained models which may lead to spurious correlations. Shape bias may also improve recognition of concepts - which is intuitive, as these samples do not contain the same textures as ImageNet samples. 

\textit{In all the models, the only regular pattern we see is that the shape bias is inversely related to ID performance. This means that as ID accuracy increases, alignment with human vision decreases.}

\subsection{Spectral Bias}
\label{subsec:spectral}
It is well known that adversarial training results in models with improved low-frequency but reduced high-frequency bias \cite{Wang_2020_CVPR,dong2019fourier,Geirhos2021_modelsvshumasn,Gavrikov_2023_CVPRW}. %
Yet, the spectral bias of other methods has been less studied, especially in correlation to other aspects of generalization beyond adversarial robustness. 

\begin{figure}
    \centering
    \begin{subfigure}{\linewidth}
        \includegraphics[width=\linewidth]{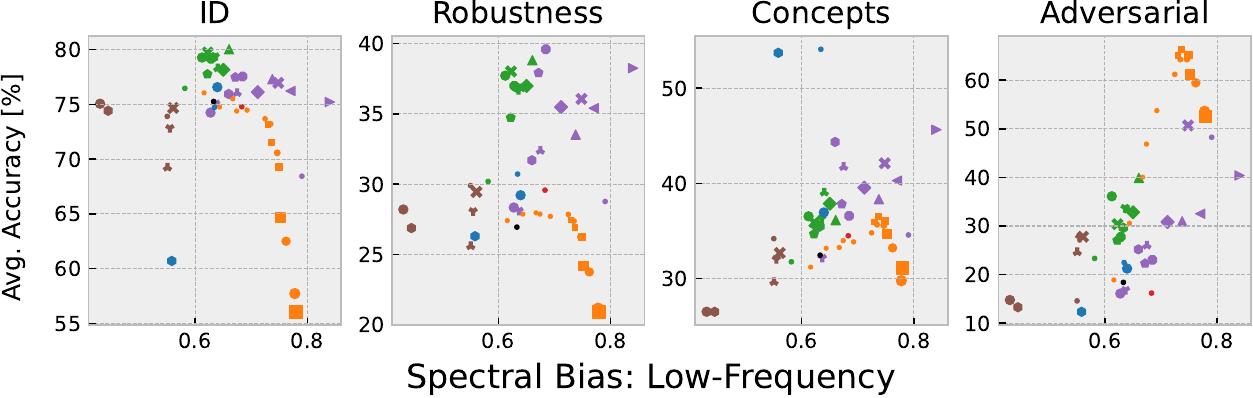}
        \vspace{-2mm}
    \end{subfigure}
    \begin{subfigure}{\linewidth}
        \includegraphics[width=\linewidth]{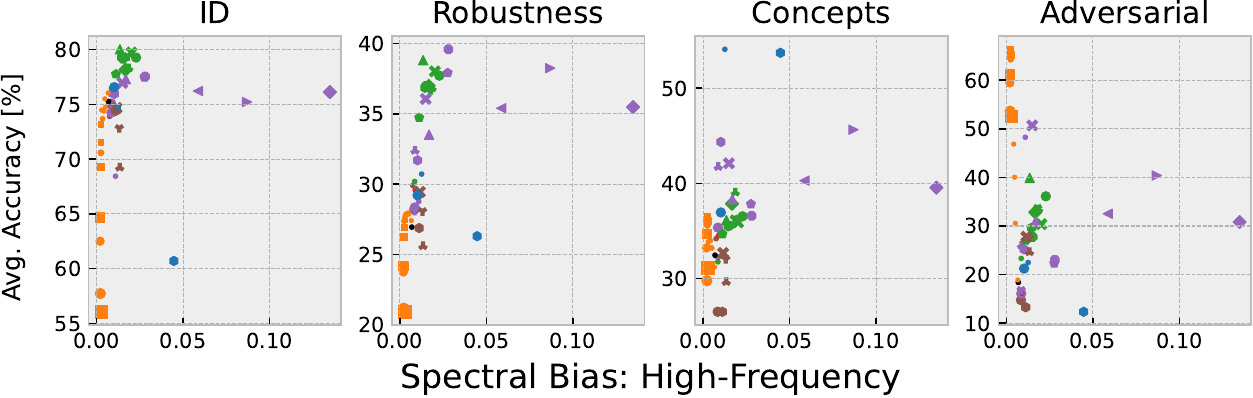}
    \end{subfigure}
    \caption{\textbf{Spectral Biases vs.~Generalization.} We correlate low-frequency (top) and high-frequency bias (bottom).}
    \label{fig:scatter_spatial}
\end{figure}

\paragraph{Low-Frequency Bias.} To our surprise, we find no meaningful correlation between a low-frequency bias and most aspects of generalization when considering all models (\cref{fig:scatter_spatial}) - except for adversarial robustness, where a stronger low-frequency bias improves adversarially robustness. This is intuitive, as attacks prefer high-frequency bands on ImageNet \cite{dong2019fourier}. Improved reliance on cues from non-affected bands is therefore a good strategy.

When separating AT from other models we again observe divergent trends. On AT, this bias shows a curvilinear correlation to all aspects of generalization (not captured by Spearman $r$ in \cref{fig:heatmaps_corr_sbs}). For concepts and adversarial robustness, some low-frequency bias helps but eventually starts hurting generalization. For ID and non-adversarial robustness, increasing low-frequency bias seems always to reduce generalization.
On non-AT models, it shows a moderate monotonic correlation with all categories except ID. However, there is often a high variance in performance. 

\textit{Improving the focus to cues from low-frequency bands improves adversarial robustness, but hardly offers any guarantees beyond that. For adversarial training, too much low-frequency bias is not desirable.} %
\paragraph{High-Frequency Bias.} A high-frequency bias, on the other hand, shows surprisingly moderate to strong correlations to all aspects of generalization (\cref{fig:heatmaps_corr_sbs}). For adversarial robustness, it is negatively correlated, but positive for all others.
Upon manual inspection of the scatterplots (\cref{fig:scatter_spatial}), we find that this correlation is slightly deceptive. There is a strong monotonic increase in generalization as models start to classify some cues in high frequency, but after some threshold, the gains stagnate or even fall. A common outlier is \textsc{ShapeNet} \cite{geirhos2018imagenettrained}, which shows some ability to recognize high-frequency cues despite being trained on data without discriminative texture cues that typically reside in these high-frequency bands. 
For adversarial robustness, we completely reject a connection due to the variance for the lowest measurements - but it seems like only models without any high-frequency bias can reach peak adversarial robustness. Unsurprisingly, AT models show next to no high-frequency bias as they are desensitized to this frequency band.

\textit{Overall, a slight capability to detect high-frequency cues seems to improve generalization except for adversarial robustness. In this context, it also seems like more high-frequency bias is better than no bias. For adversarial robustness, even the slightest increase in high-frequency bias decreases performance rapidly.}

\begin{figure}
    \centering
    \begin{subfigure}{\linewidth}
        \includegraphics[width=\linewidth]{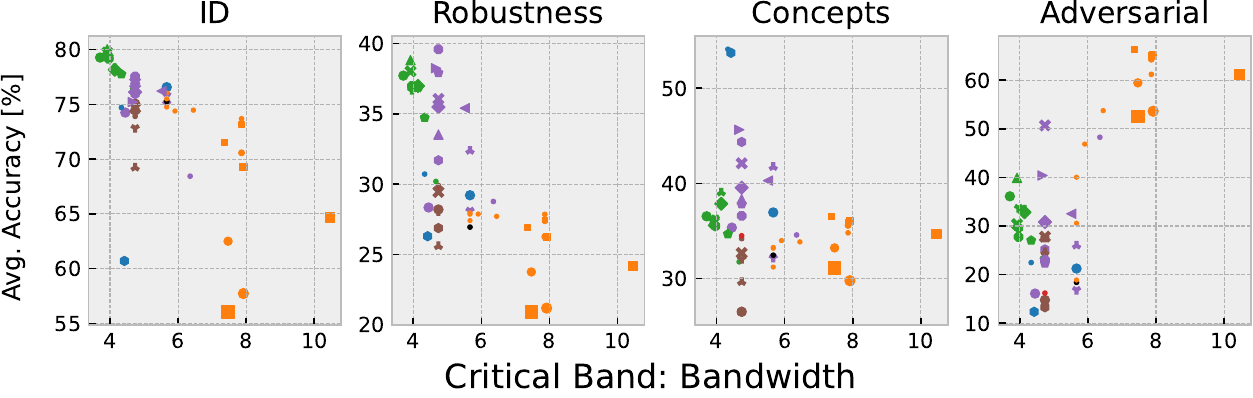}
        \vspace{-2mm}
    \end{subfigure}
    \begin{subfigure}{\linewidth}
        \includegraphics[width=\linewidth]{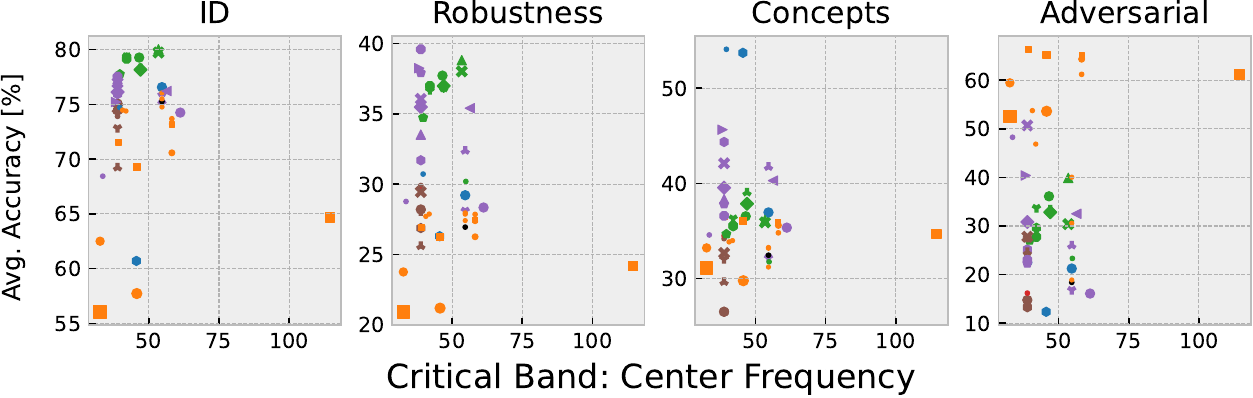}
        \vspace{-2mm}
    \end{subfigure}
    \begin{subfigure}{\linewidth}
        \includegraphics[width=\linewidth]{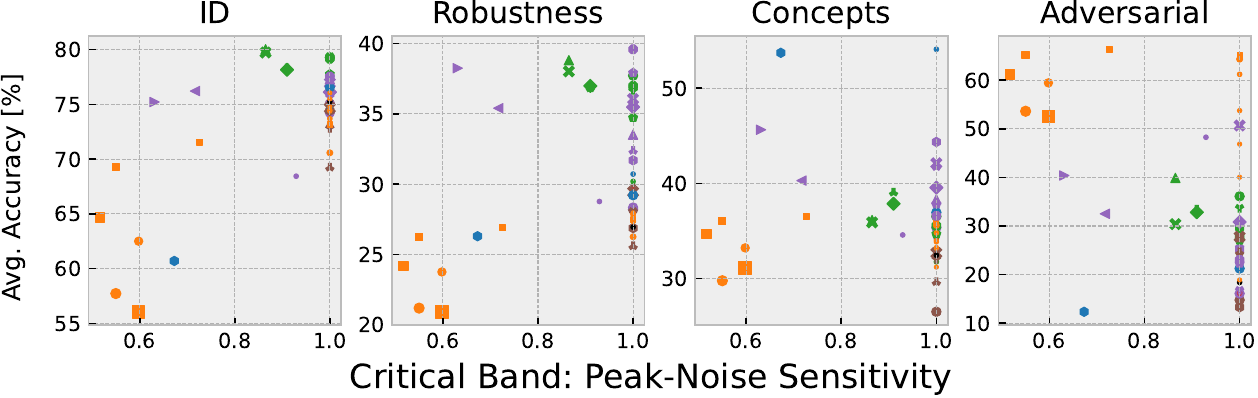}
    \end{subfigure}
    \caption{\textbf{Critical Band vs.~Generalization.} We correlate three properties: the bandwidth (top), center frequency (center), and peak-noise sensitivity (bottom).}
    \label{fig:scatter_cb}
\end{figure}
\subsection{Critical Band}
\label{subsec:criticalband}
The critical band parameterized by its bandwidth, center frequency, and peak-noise sensitivity has been recently shown to correlate with adversarial robustness (and shape bias) \cite{subramanian2023spatialfrequency}. Akin to what we have observed on other biases, results show different trends between adversarially trained (AT) and other models. For AT, a higher bandwidth of the critical band correlated positively with adversarial robustness, but the correlation was negative for other models. We repeat the measurements using a more careful collection of models and benchmarks. Additionally, we tweak the original test by using normalization, more samples, and more classes. Our normalization allows us to include models trained under $\ell_\infty$-norm and larger attack budgets ($\epsilon$) that couldn't be tested with the original methodology and allows us to scale the evaluation against all 1,000 classes. In both cases, unnormalized measurements will break the detection of the critical band as all regions would be classified as critical. Overall, our measurements are more accurate, but may not be directly comparable to the original study. \cref{app_sec:criticalband} contains results using the original methodology (and in combination with normalization).

\paragraph{Bandwidth.} The bandwidth not only moderately correlates with adversarial robustness (similarly to the findings in \cite{subramanian2023spatialfrequency}), but also strongly correlates with ID and robustness (\cref{fig:heatmaps_corr_sbs}). However, there is an inverted trend and the correlation to adversarial robustness is negative. For generalization to concepts we only see a weak positive correlation. Generally, we observe many outliers in the scatterplots (\cref{fig:scatter_cb}) and often the best performance is not achieved by the models with the lowest bandwidth. Further, some trends only become noticeable or accelerated by adversarially-trained models which accumulate for models with the lowest bandwidth. It remains to be discovered whether low-critical-bandwidth non-AT ResNet-50s exist and how they perform on generalization tests.

\paragraph{Center Frequency.} In contrast to the original study, we find \textit{no correlation} between the center frequency and any of our benchmarks (\cref{fig:scatter_cb}). Even when we single out adversarially trained models, we still observe no statistically significant correlation.

\paragraph{Peak-Noise Sensitivity.} We measure a slight statistical correlation among both adversarial training (AT) and non-AT models (\cref{fig:heatmaps_corr_sbs}). However, the significant concentration (\cref{fig:scatter_cb}) around the value of $1$ indicates an issue in the test and suggests that \textit{this metric might not effectively capture trends in extensive evaluations such as ours}. As for the rest, AT models don't exhibit a distinct clustering pattern, yet display a notably strong positive correlation with both ID and non-adversarial robustness. 

\textit{For the most part, the critical band seems to poorly explain generalization. However, a lower bandwidth seems to be necessary for generalization except for adversarial robustness where the opposite holds.}

\section{Other observations}
\label{sec:other_observations}
We also notice a couple of other observations that may not directly fit into our storyline but may still be interesting.

\paragraph{Improved low-frequency performance does not guarantee a shape bias.} A shape bias causes a model to prefer global over local information as textures are mostly local information found in higher frequency bands. Indeed, we find a moderately strong positive correlation between low-frequency and shape bias ($r=0.66, p=2e-4$), but also clear outliers in both directions: \textsc{ShapeNet} (trained on SIN) \cite{geirhos2018imagenettrained} achieves a high shape bias of $0.81$ but only a low-frequency bias of $0.56$; the baseline \cite{resnet} achieves a low-frequency bias of $0.63$ but only a shape bias of $0.21$. %

\paragraph{Scaling data does not guarantee a shape bias.} Previous works have observed a trend where upscaling (pre-)training data results in higher shape bias (and better generalization) \cite{Geirhos2021_modelsvshumasn,dehghani23a}. However, when comparing a ResNet-50 trained with self-supervised learning \cite{yalniz2019billionscale} on \textit{YFCC-100M} \cite{yfcc} containing 100 million additional samples, we notice that it obtains the same $0.21$ shape bias as the baseline despite generally better performance. This shows that the shape bias is not simply determined by the dataset scale alone.

\paragraph{Performance on \cbar~correlates well with most other benchmarks.} Analogous to \cite{NEURIPS2020_d8330f85}, we notice that ImageNet accuracy strongly correlates with the average ($\overline{r}=0.79$) of the other benchmarks (excluding SIN and PGD) (\cref{fig:heatmap_benchmarks_no_sslswsl}) which suggest that ImageNet performance is key to most aspects of generalization.  However, we notice that \cbar~shows an even stronger correlation with the remaining benchmarks ($\overline{r}=0.83$). Even its weakest correlation (IN-A) is still moderately strong. Yet, just like ImageNet, it does not correlate with PGD and SIN. Future studies may prefer this benchmark to study robustness if they are restricted in their compute budget - \cbar~is a faster benchmark than IN-C due to fewer corruption classes.

\begin{figure}
    \centering
    \includegraphics[width=\linewidth]{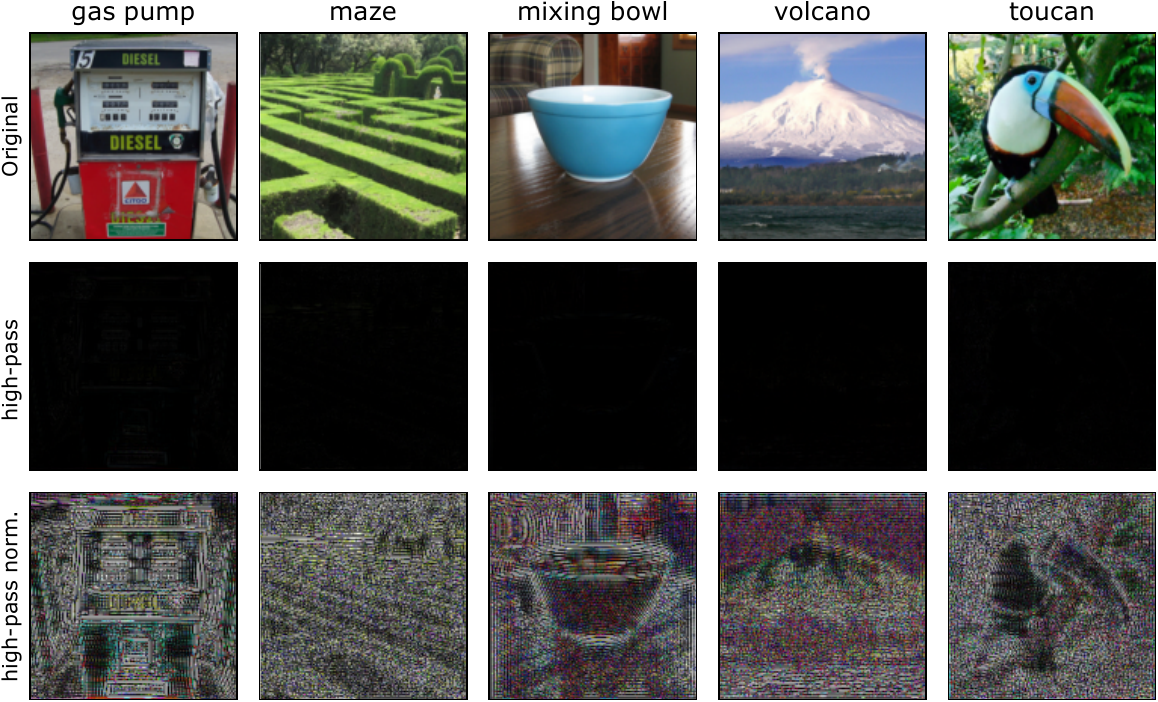}
    \caption{\textbf{Examples of high-pass (30\%) filtered ImageNet samples.} High-pass filtered samples (center) are correctly classified by \textsc{DeepAugment + AugMix} \cite{hendrycks2021rendition} but not by most other models and unlikely by humans. Only \textit{std} normalized samples (bottom) reveal some similarity to the original samples (top).}
    \label{fig:imagenet_hfc}
\end{figure}
\section{Discussion and Conclusion}
\label{sec:discussion}

We have identified numerous shortcomings in previous studies showing correlations between shape bias, spectral biases, and the critical band to generalization. Our holistic and more carefully designed study shows that while correlations between some of these biases and \textit{some} benchmarks exist, most of them fail to correlate to a broader view of generalization. 

This leaves us with the question \textit{Is it useful to directly optimize for these biases?} \Ie, should we use these biases as priors?
Ideally, a prior would \textit{monotonically} and \textit{causally} correlate with all aspects of generalization. However, for the tested biases, we have seen many outliers, non-monotonic trends, and/or only correlations with some aspects of generalization. It is important to understand, that in this case the utility of the bias concerning holistic generalization is significantly lowered, but it does not necessarily mean that the bias itself is not useful. Instead, it only tells us that such a bias alone is not sufficient to improve generalization but might be still helpful in combination with other aspects. Beyond that, there might be other valid reasons (\eg, interpretability or alignment) to optimize these biases.

Additionally, we have seen that most biases show strong correlations on adversarially-trained models (\eg, shape bias, low-frequency bias), but not on the entire model zoo. On the one hand, this means that the bias is not useful independent of the training mode, \ie, it does not correlate if we discard this metadata. On the other hand, this might also be a limitation of our study as our AT models are extremely homogenous compared to the other models. All of the models were trained with the same hyperparameters except for the adversary parameters (see \cite{salman2020transfer}). While we cannot reject causal correlations based on our model zoo, we cannot prove them either and look forward to future studies that study adversarial training more closely.

Finally, we have seen some counter-intuitive trends where models misaligned with human perception perform better - \eg, a stronger texture bias increases ID performance and a high-frequency bias seems to improve generalization (except for adversarial robustness) despite containing next to no discriminative cues (see \cref{fig:imagenet_hfc} for examples). For the latter, it is worth noting that even the strongest high-frequency biased model only detects an average of $9\%$ of the samples based on this band alone. As such, we cannot claim that a high-frequency bias is all you need, but drawing some predictive cues from this band seems to be reasonable. In general, we want to issue a word of caution as we are concerned that the reason for improvements we have seen might be due to dataset biases, \eg frequency shortcuts \cite{Wang_2023_ICCV} and not an intrinsically better representation. A counter-argument to that theory is that we have seen correlations on many semantically different datasets and it seems unlikely that they all contain the same shortcuts. Yet, we are curious to see if future work can derive a better understanding of generalization and design better benchmarks.

In summary, it seems that the generalization in neural networks - even when fixing the architecture - is too complex to be explained by a single bias. However, certain biases may be necessary (\eg, a low bandwidth of the critical band, or a bit of high-frequency bias) to achieve generalization. It remains to be shown if a single bias or a combination of them exists that can explain generalization better.

\section*{Acknowledgements}
We thank Robert Geirhos and Ross Wightman for their help regarding model details and Ajay Subramanian for his help in replicating the initial study results.

\newpage
{
    \small
    \bibliographystyle{ieeenat_fullname}
    \bibliography{main}
}

\clearpage
\crefalias{section}{appendix}
\renewcommand{\thesection}{\Alph{section}}
\setcounter{section}{0}
\maketitlesupplementary

\renewcommand\contentsname{} %

\begingroup
\let\clearpage\relax
\vspace{-1cm} %

\startcontents[appendix]
\printcontents[appendix]{l}{1}{\section*{Overview}\setcounter{tocdepth}{2}}
\endgroup

\section{Broader Impact}

This study underscores the critical importance of conducting experiments rigorously to derive valid conclusions. By addressing the limitations in prior research and cautioning against overreliance on bias measures without comprehensive validation, this work highlights the necessity of robust experimental methodologies to advance our understanding of neural network generalization.
It is well understood that correlation does not prove causality. While this study mainly challenges established correlations for a more nuanced understanding of the influences of biases on generalization, it also presents new ones (\eg, high-frequency bias) - it is important that future works equally rigorously evaluate our findings and try to break the causality.

\paragraph{Transferability and Domain-Specific Considerations.}
This study explores biases in the ImageNet classification problem. While this dataset and problem are representative of a significant portion of computer vision research, our (and previous) findings may be limited in their transferability to other problems. For example, while there is a theoretical grounding for shape bias in object recognition, it is not intuitively clear if this applies to medical classification tasks such as melanoma detection.
We ask researchers to exercise caution when extrapolating findings from specific contexts to broader applications and encourage a rigorous evaluation of model performance on their specific problem - in particular, in safety-critical domains where human lives are at stake.

\section{Tables of Results}
\label{app_sec:model_tables}
\Cref{supp_tab:performance} contains an overview of all our models with detailed performance on every benchmark (\ie, all datasets and the adversarial attack). It also serves as a legend for the markers in all our scatter plots (main paper and Appendix). \Cref{sup_tab:influences} contains the corresponding measurements of our studied biases.

\section{Details about Generalization Benchmarks}
\label{app_sec:dataset_details}

\paragraph{In distribution (ID).}
\begin{itemize}
    \item The \textit{ImageNet (IN)} \cite{imagenet} validation set is the standard test dataset, containing 50,000 images with 1,000 different classes. 
    \item \textit{ImageNet v2 (IN-v2)} \cite{recht19imagenet2} is a newer 10,000 images test set sampled a decade later following the methodology of the original curation routine. 
    \item \textit{ImageNet-ReaL (IN-ReaL)} \cite{beyer2020imagenet} is a re-annotated version of the original ImageNet validation set. It assigns multiple labels per image and contains multiple corrections of the original annotations.
\end{itemize}

\paragraph{Robustness.} 
\begin{itemize}
    \item \textit{ImageNet-C (IN-C)} \cite{hendrycks2019commoncorruptions} is a dataset consisting of 19 synthetic corruptions applied to the original ImageNet validation set under increasing severity. The original protocol suggests averaging the error over all corruptions and severities normalized by the error of AlexNet \cite{alexnet} on the same. Contrary, we simply report the mean accuracy over all 19 corruptions and severity levels\footnote{The results remain comparable by a simple linear transformation}. 
    \item \textit{ImageNet-\={C} (\cbar)} \cite{mintun2021cbar} extends IN-C by adding 10 new corruptions which were chosen to be perceptually dissimilar but conceptually similar to the corruptions in IN-C. We report the mean accuracy akin to IN-C.
    \item \textit{ImageNet-A (IN-A)} \cite{hendrycks2021nae} contains 7,500 additional images belonging to 200 ImageNet classes that are naturally hard to classify for ImageNet models and are, thus, posing natural adversarial examples.
\end{itemize}

\paragraph{Concepts.}
\begin{itemize}
    \item \textit{ImageNet-Renditions (IN-R)} \cite{hendrycks2021rendition} is a dataset of 30,000 images of 200 different ImageNet classes in different styles, such as cartoons, paintings, toys, etc.
    \item \textit{ImageNet-Sketch (IN-S)} \cite{wang2019learning} is a dataset of over 50,000 hand-drawn sketches belonging to all ImageNet classes. Semantically it can be seen as a subset of IN-R.
    \item \textit{Stylized ImageNet (SIN) }\cite{geirhos2018imagenettrained} contains ImageNet validation images that have been stylized using different artistic filters to destroy texture information. We use the official 16-class subset given in \cite{Geirhos2021_modelsvshumasn}.
\end{itemize}

\paragraph{Adversarial Robustness.} We use a Project Gradient Descent (PGD) \cite{madry2018towards} attack to adaptively evaluate robustness. As models trained without adversarial training \cite{madry2018towards} are highly susceptible to such attacks, we attack with a reduced budget of $\epsilon=0.5/255$ under $\ell_\infty$ norm, using 40-steps with $\alpha=2/255$. This benchmark is an important data point due to its adaptive nature. While in theory, a model may overfit a static dataset due to a finite number of test samples, (ideal) adaptive benchmarks would not affected. 

\section{Results on ViT}
\label{app_sec:vit}

\begin{figure}
    \centering
    \begin{subfigure}{\linewidth}
        \includegraphics[width=\linewidth]{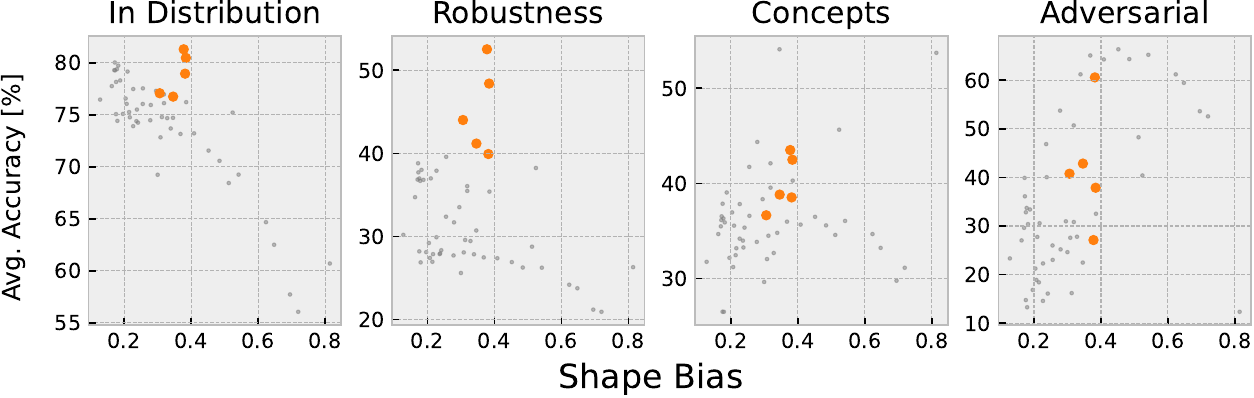}
        \vspace{-3mm}
    \end{subfigure}
    \begin{subfigure}{\linewidth}
        \includegraphics[width=\linewidth]{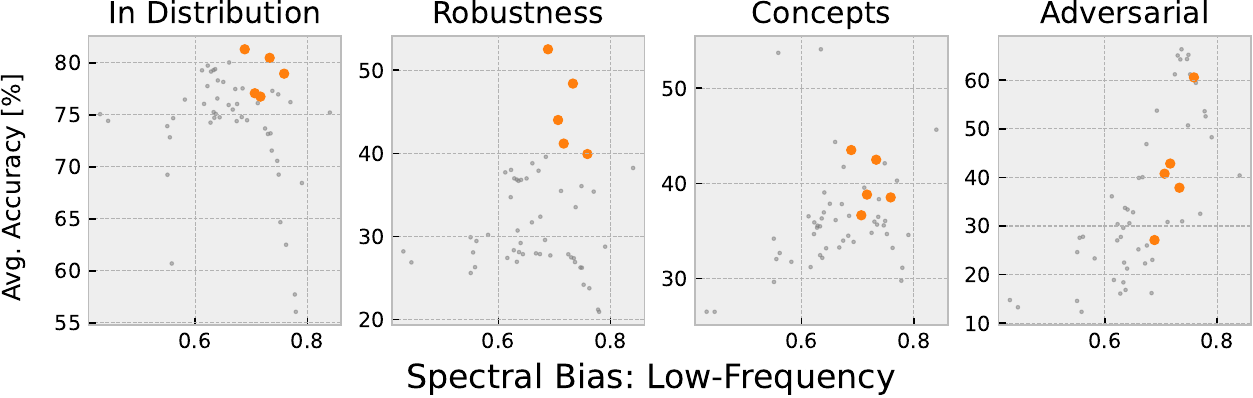}
        \vspace{-3mm}
    \end{subfigure}
    \begin{subfigure}{\linewidth}
        \includegraphics[width=\linewidth]{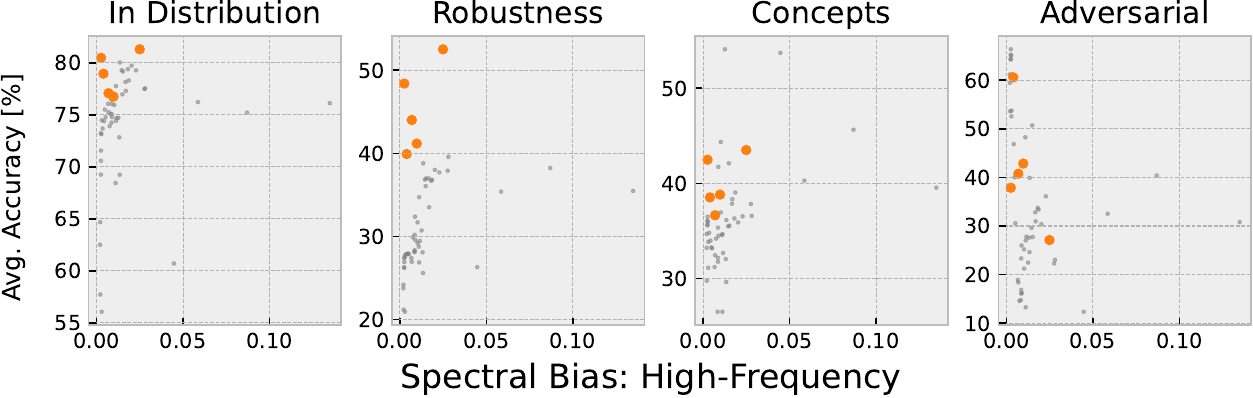}
        \vspace{-3mm}
    \end{subfigure}
    \begin{subfigure}{\linewidth}
        \includegraphics[width=\linewidth]{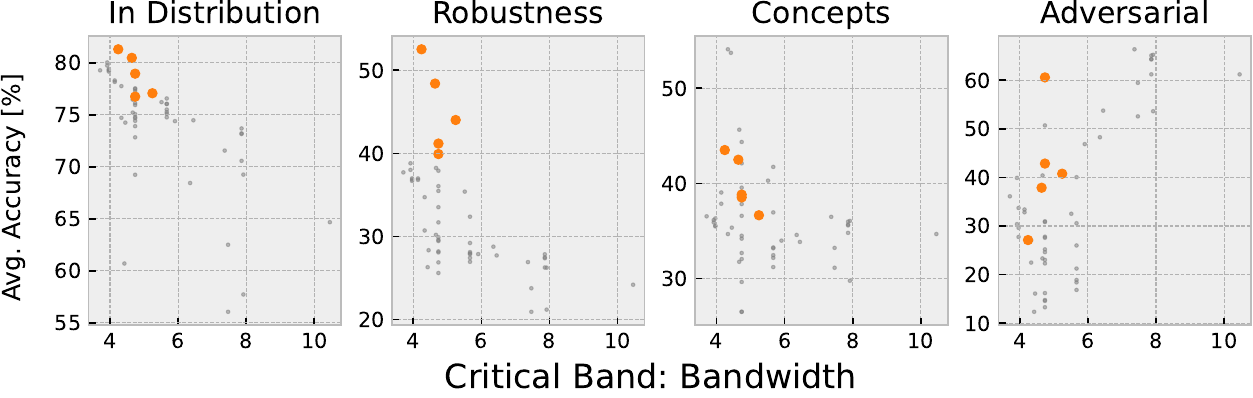}
        \vspace{-3mm}
    \end{subfigure}
    \begin{subfigure}{\linewidth}
        \includegraphics[width=\linewidth]{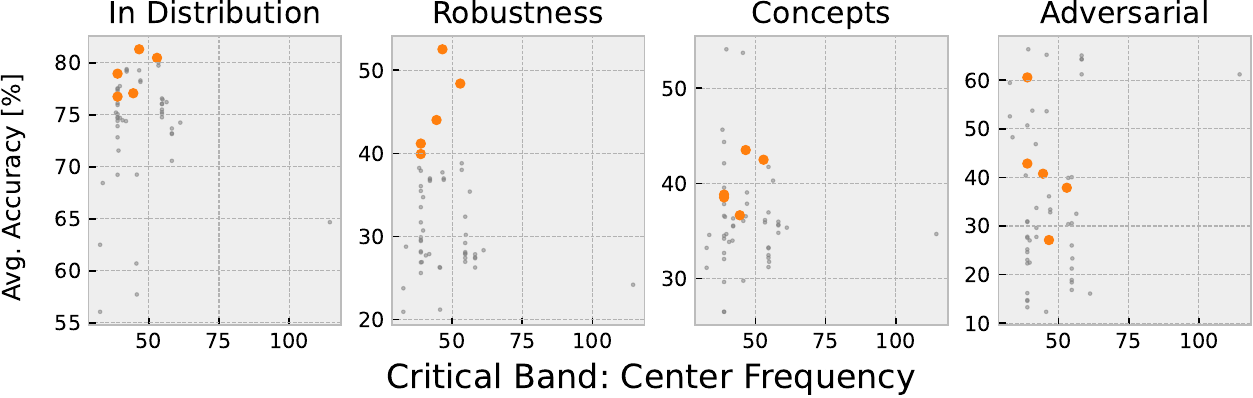}
        \vspace{-3mm}
    \end{subfigure}
    \begin{subfigure}{\linewidth}
        \includegraphics[width=\linewidth]{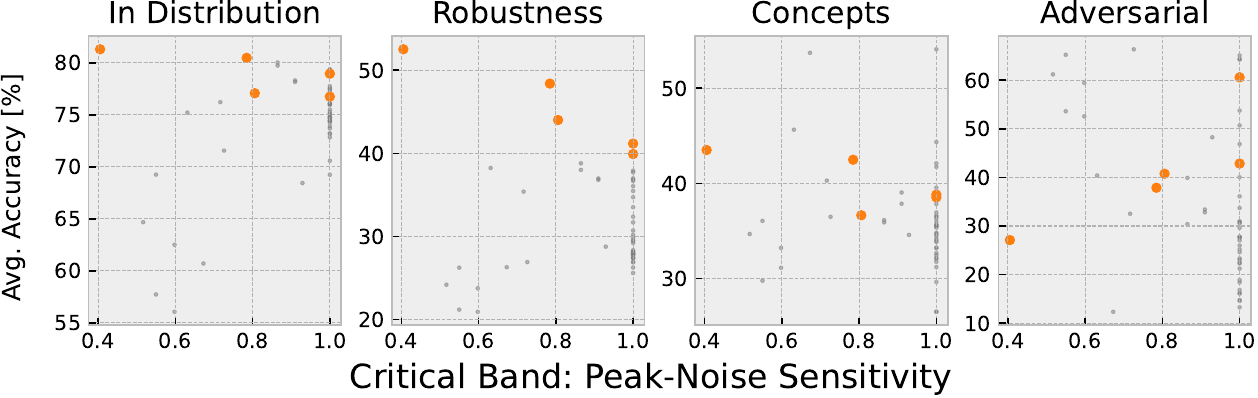}
    \end{subfigure}
    \caption{Biases vs.~Generalization on ViT-B/16-based models. Orange markers represent the ViTs, gray ones are ResNet-50s.}
    \label{supp_fig:vits}
\end{figure}

In line with our results on ResNet-50, we provide results on ViT-B/16 \cite{dosovitskiy2021an} in \cref{supp_fig:vits}. Unlike most ViTs, these models are exclusively trained on the ILSVRC2012 subset of ImageNet. The models originate from AugReg \cite{steiner2022how}, Masked Autoencoders (MAE) \cite{he2022mae}, DINO \cite{caron2021emerging}, data-efficient image transformers (DeiT) \cite{touvron21a}, and sharpness-aware minimizers (SAM) \cite{chen2022when}.

\section{Detailed Plots for the Analysis}
\label{app_sec:detailed_plots}

\begin{figure}[!h]
    \centering
    \includegraphics[width=\columnwidth]{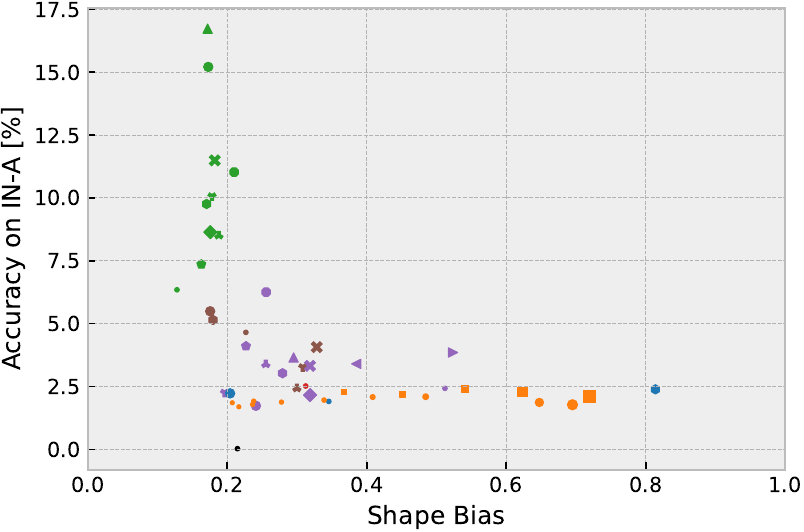}
    \caption{\textbf{Shape Bias vs.~IN-A.} Only strongly texture-biased models show significant improvements but the most texture-biased model is not the best IN-A model. Markers indicate models as described by the legend in \cref{supp_tab:performance}.}
    \label{supp_fig:ina_shapebias}
\end{figure}

In \cref{subsec:shape_bias} we discuss the correlation between the performance on IN-A and the shape bias. We provide the plot for this in \cref{supp_fig:ina_shapebias}. As discussed, only strong texture-biased models show improvements in IN-A performance. 

\begin{figure}[!h]
    \centering
    \begin{subfigure}{\linewidth}
        \includegraphics[width=\linewidth]{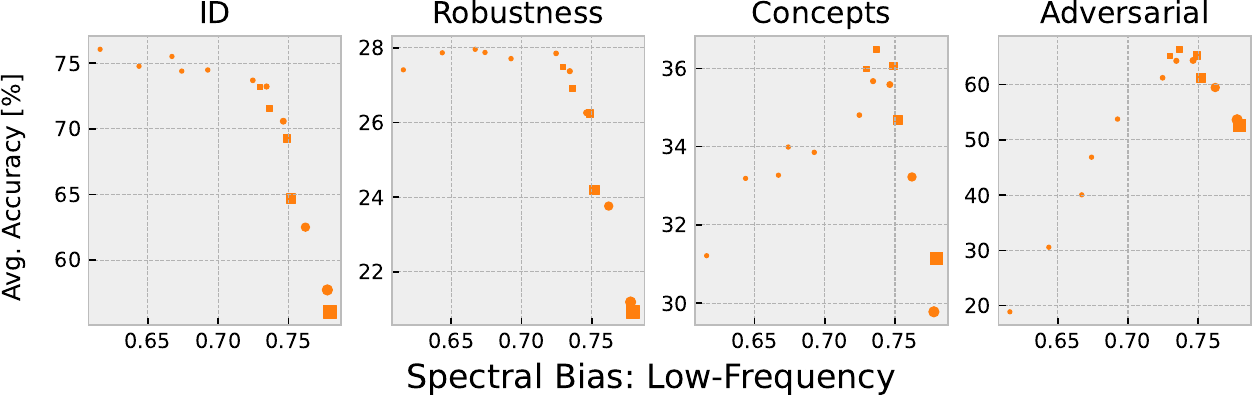}
        \vspace{-3mm}
    \end{subfigure}
    \begin{subfigure}{\linewidth}
        \includegraphics[width=\linewidth]{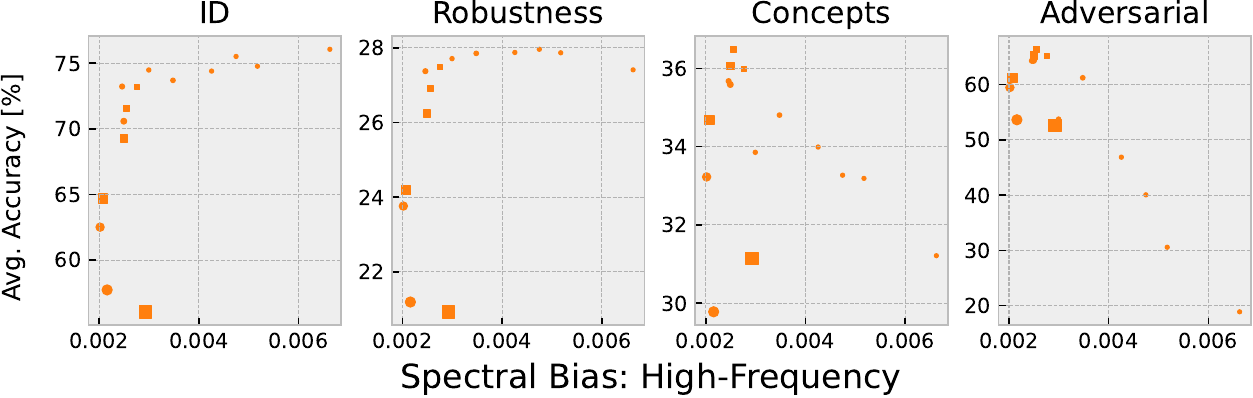}
    \end{subfigure}
    \caption{\textbf{Spectral Biases vs.~Generalization only on adversarially-trained (AT) models.} Markers indicate models as described by the legend in \cref{supp_tab:performance}. Marker size correlates with the attack budget $\epsilon$ during training.}
    \label{supp_fig:spectal_indicators_for_at}
\end{figure}

Additionally, we discuss in \cref{subsec:spectral} the non-causal - and in particular non-functional/non-injective - relationship between high-frequency bias and generalization for AT models. In \cref{supp_fig:spectal_indicators_for_at} we show the same results as in the main paper (\cref{fig:scatter_spatial}) but limited to AT models to show the trend more clearly. Also, note how there is also a non-functional/non-injective relationship to the attack budget $\epsilon$ (corresponding to the marker size).

\begin{figure}[!h]
    \centering
    \includegraphics[width=\columnwidth]{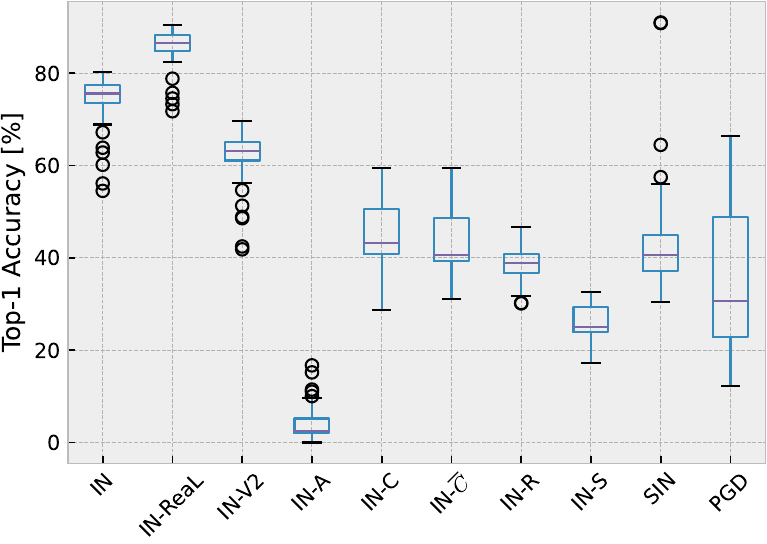}
    \caption{\textbf{Performance distribution} of the model zoo on all individual benchmarks.}
    \label{supp_fig:boxplot_benchmarks_accuracy}
\end{figure}

\begin{figure}[!h]
    \centering
    \includegraphics[width=\columnwidth]{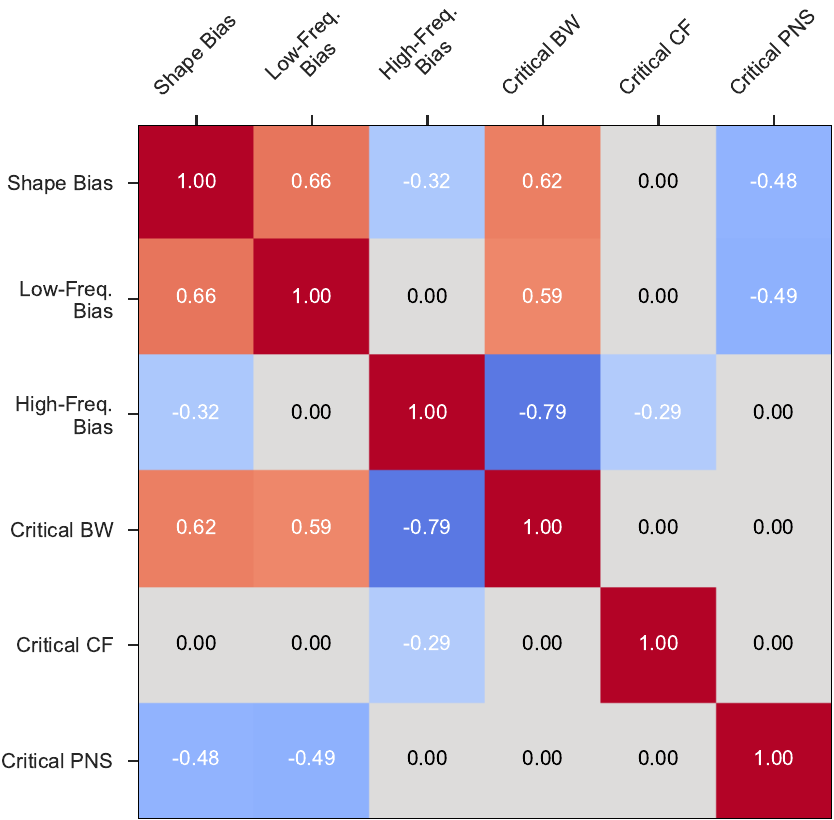}
    \caption{\textbf{Correlations between biases.} Correlations measured by Spearman $r$. We set non-significant correlations with $p\geq 0.05$ to 0.}
    \label{supp_fig:heatmap_indicators}
\end{figure}

\begin{figure*}
    \centering
    \includegraphics[width=\textwidth]{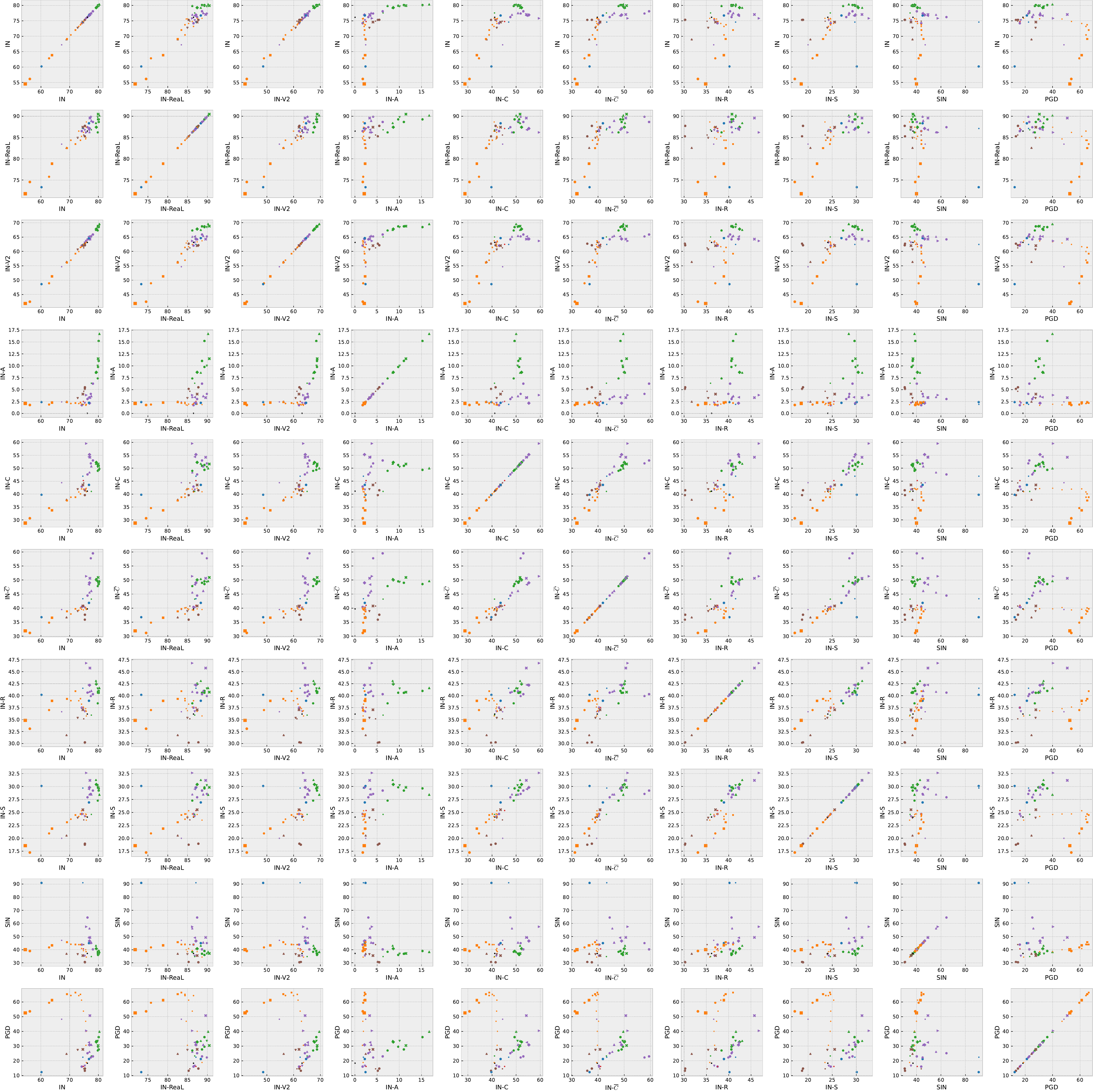}
    \caption{Performance comparison on all dataset pairs. Markers indicate models as described by the legend in \cref{supp_tab:performance}.}
    \label{supp_fig:scatter_benchmarks}
\end{figure*}

We also discuss the statistical correlation between our benchmarks in the main paper (\cref{sec:prev_results}). In \cref{supp_fig:scatter_benchmarks} we additionally provide scatter plots between all benchmark pairs. We also show the distribution of reached accuracy on all benchmarks in \cref{supp_fig:boxplot_benchmarks_accuracy}, and provide an overview of the statistical correlation between our biases in \cref{supp_fig:heatmap_indicators} which we use to discuss the relationship between shape bias and spectral biases in \cref{sec:other_observations}.
\section{Changes to the Critical Band Test}
\label{app_sec:criticalband}

A cornerstone of the initial study \cite{subramanian2023spatialfrequency} is the evaluation of classification accuracy on noise-modified subsets of ImageNet. Each subset contains on average only 30 random ImageNet samples that are gray-scaled, contrast-reduced, and introduced to Gaussian noise at specified frequencies and varying intensities. In contrast, we use all 50,000 ImageNet samples for each subset (and apply the same transformations). This way, we get less noisy results by testing more samples and making sure that all subsets contain the exact same images.
Then, the original test measures the accuracy against 16 super-classes for each subset. This was mainly done for comparability reasons for the same study with the human subjects. Since we do not compare to human trials, we measure the common top-1 accuracy against all 1,000 classes.
Finally, the critical band is measured by a fitted Gaussian function to the resulting heatmaps. The authors binarized the heatmaps by performance with a threshold of 50\%. As we discussed in \cref{sec:method} this is not ideal for models that are contrast-sensitive (\eg, AT models) and does not allow evaluation of such models. Thus, we normalize the heatmap by the maximum accuracy over all tests prior to the curve fitting. Under our methodology, this corresponds to normalization with the accuracy on the contrast-reduced images (but not under the original test where the random samples for each subset introduce noise).

\begin{figure*}
    \centering
    \begin{subfigure}{0.48\linewidth}
        \begin{subfigure}{\linewidth}
            \includegraphics[width=\linewidth]{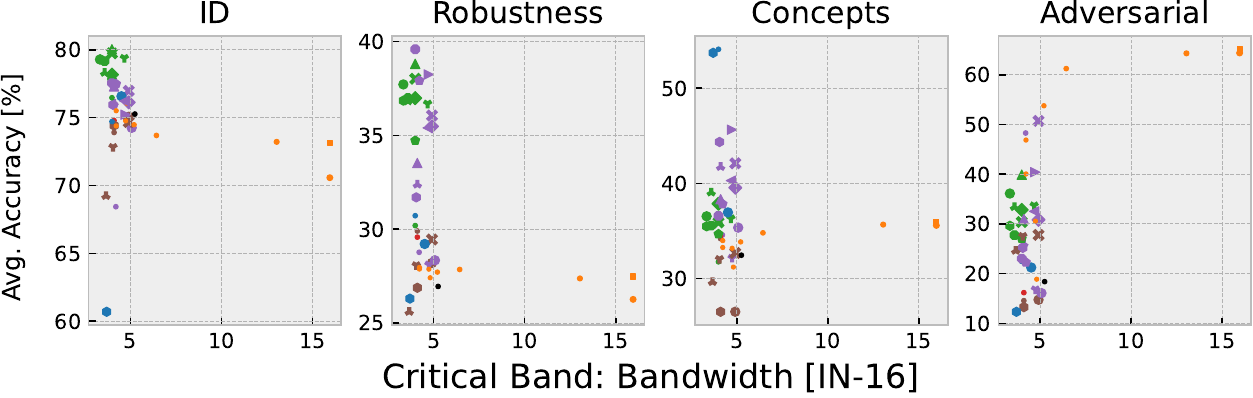}
        \end{subfigure}%
        \vspace{0.1in}
        \begin{subfigure}{\linewidth}
            \includegraphics[width=\linewidth]{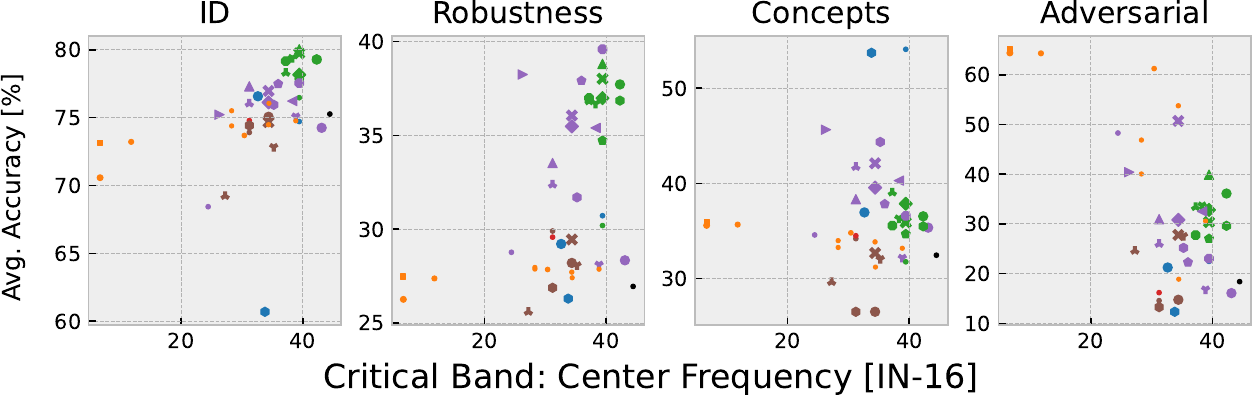}
        \end{subfigure}%
        \vspace{0.1in}
        \begin{subfigure}{\linewidth}
            \includegraphics[width=\linewidth]{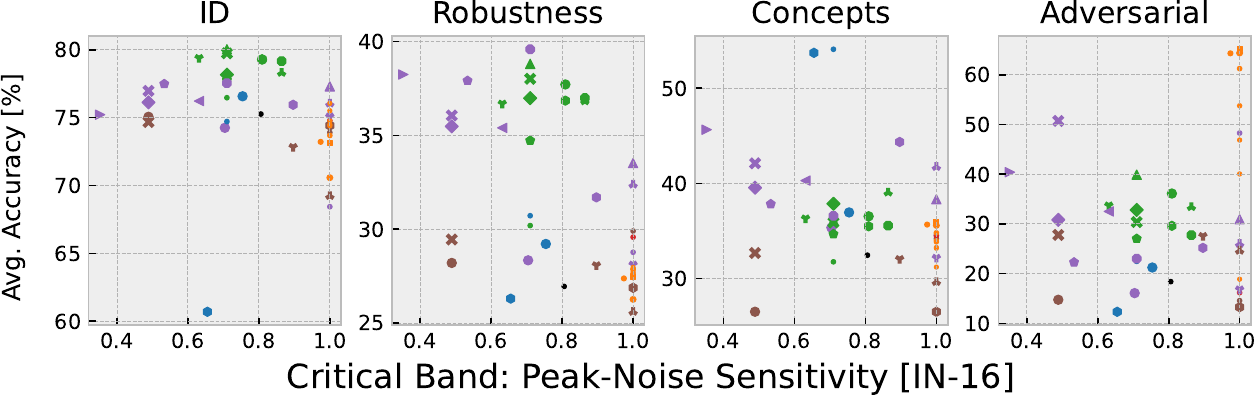}
        \end{subfigure}%
        \vspace{0.1in}
        \caption{\textbf{Original test on IN-16.}}
        \label{supp_fig:sub_critical16_orig}
    \end{subfigure}%
    \hfill
    \begin{subfigure}{0.48\linewidth}
        \begin{subfigure}{\linewidth}
            \includegraphics[width=\linewidth]{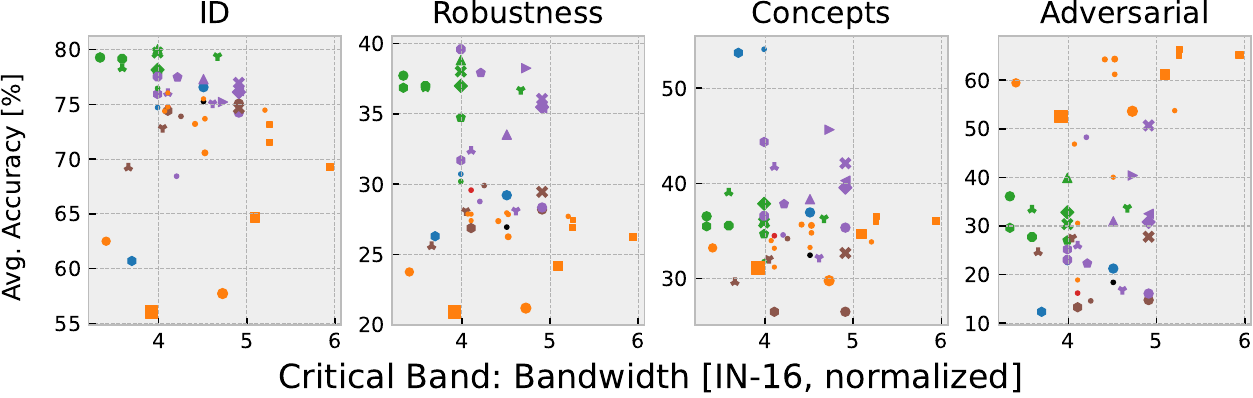}
        \end{subfigure}%
        \vspace{0.1in}
        \begin{subfigure}{\linewidth}
            \includegraphics[width=\linewidth]{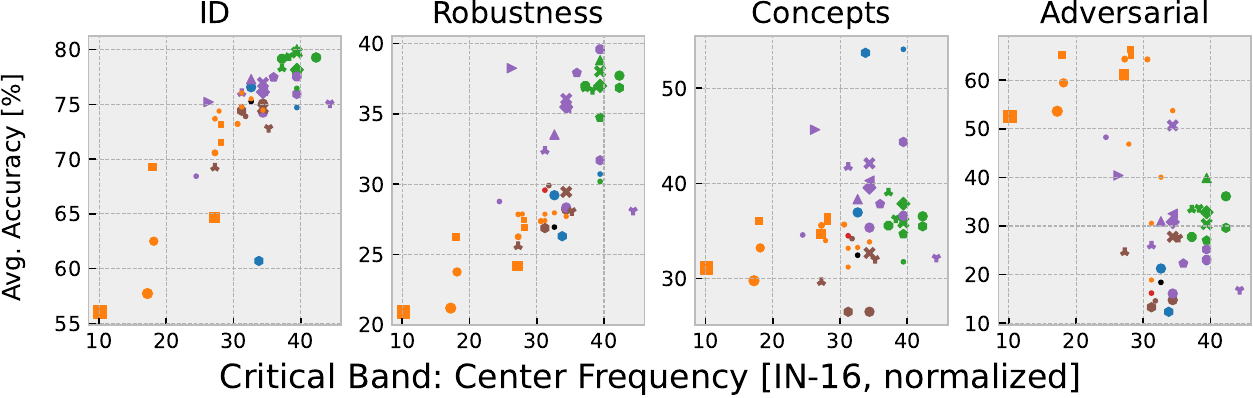}
        \end{subfigure}%
        \vspace{0.1in}
        \begin{subfigure}{\linewidth}
            \includegraphics[width=\linewidth]{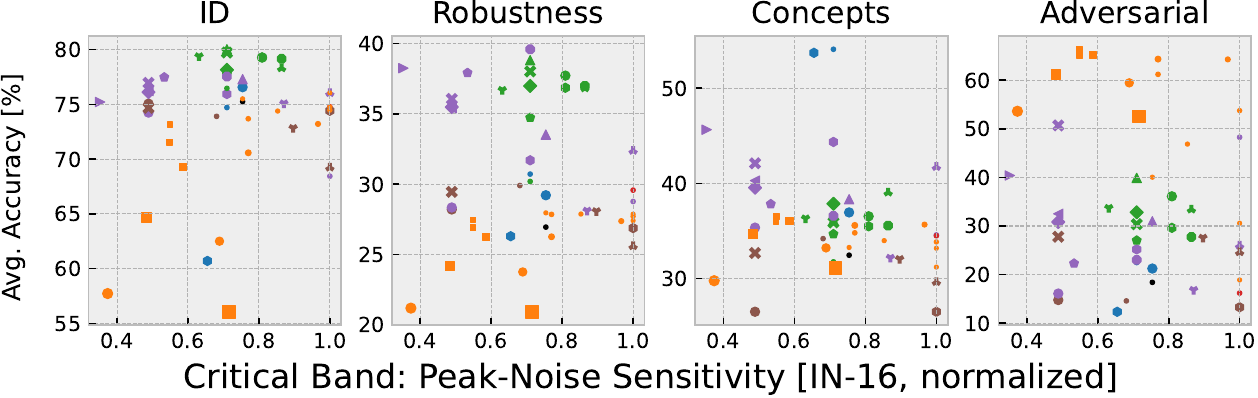}
        \end{subfigure}%
        \vspace{0.1in}
        \caption{\textbf{Original test on IN-16 with normalization.}}
        \label{supp_fig:sub_critical16_norm}
    \end{subfigure}%
    \vspace{0.2in}
    \begin{subfigure}{\linewidth}
        \includegraphics[width=\linewidth]{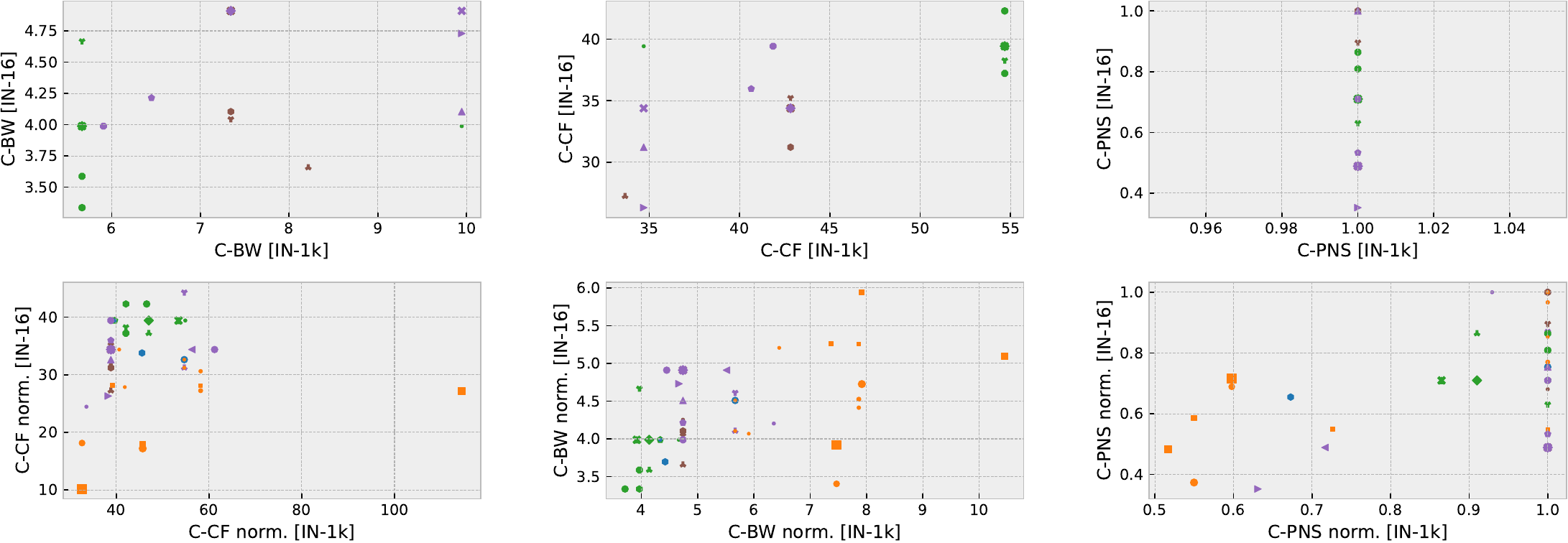}\vspace{0.1in}
        \caption{\textbf{Critical band evaluation on IN-1k and IN-16 in comparison.} We compare original (top) and normalized (bottom) evaluations.}%
        \label{supp_fig:sub_critical16_v_in1k}
    \end{subfigure}
    \caption{Measurement of the critical band following the original methodology of \citet{subramanian2023spatialfrequency}. \textbf{(a)} original test; \textbf{(b)} original test with normalized accuracy; \textbf{(c)} Comparison between results in ImageNet  (IN-1k) as in the main paper and the 16-super-class subset (IN-16). Models with unreasonable measurements ($\text{C-BW} \geq 100$) were removed. Markers indicate models as described by the legend in \cref{supp_tab:performance}.}
    \label{supp_fig:critical16}
\end{figure*}

For completeness, we have also evaluated our model zoo with the original method and the results are shown in \cref{supp_fig:critical16}. Additionally, to the exact same evaluation (\cref{supp_fig:sub_critical16_orig}) we also apply normalization (\cref{supp_fig:sub_critical16_norm}). Under the original condition, we see no reasonable correlation for any bias except when limiting the study to AT models. For the normalized study, we again see no correlations between any bias for all models, except for ID and Robustness which show some correlation to the center frequency. However, the correlation is mostly determined by the tail of AT models - removing these models would break the correlation and, thus, make a causal connection highly unlikely. 

In \cref{supp_fig:sub_critical16_v_in1k} we show the scatter plots between IN-1k and IN-16 obtained results and do not see a correlation indicating that the results obtained by our method deviate from the original test. While our modifications may not be perfect either (\eg, both our and \citet{subramanian2023spatialfrequency} arbitrarily pick 50\% as threshold) our modifications are theoretically grounded and, thus, introduce an improved measurement of the critical band for models.

\section{General Observations on the Low/High-Pass Data}
\label{app_sec:bandpass}
\begin{figure}[h]
    \centering
    \includegraphics[width=\linewidth]{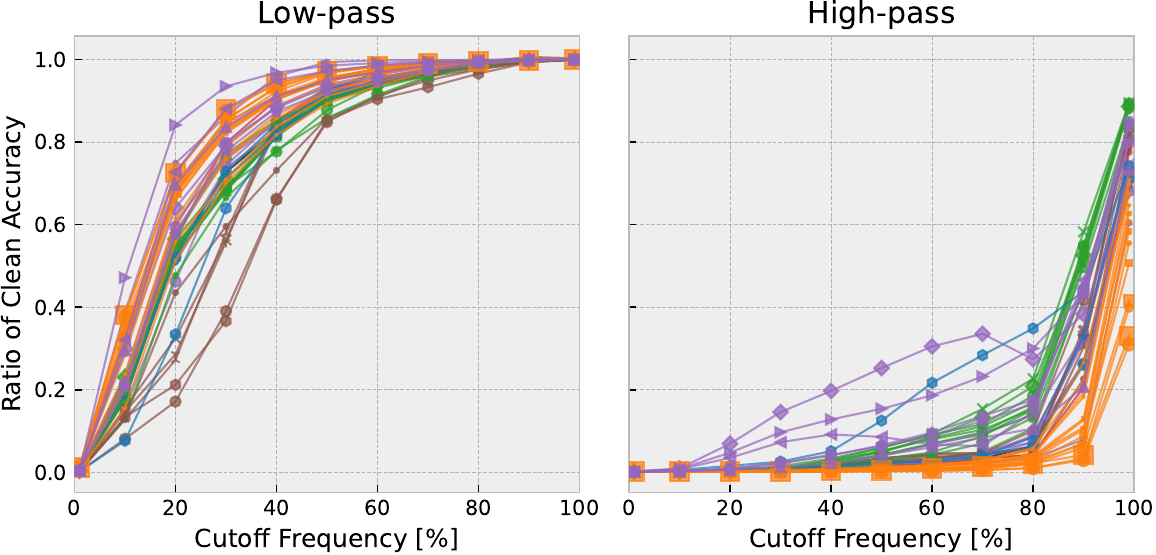}
    \caption{Frequency band-pass test on ImageNet accuracy using low-pass (left) and high-pass (right) filters with increasing cutoff frequency. The distance to the original image decreases with increasing cutoff.}
    \label{fig:bandpass}
\end{figure}

In this section, we want to provide some high-level findings on our low/high-pass data test (based on \cref{fig:bandpass}).
Contrastive learning models underperformed the baseline in low-frequency bias but performed on par for high-frequency bias. We cannot prove this to be indicative of a shortcoming of contrastive learning, as we primarily benchmark older techniques that perform worse than supervised learning because newer methods are almost exclusively designed for ViTs. Still, this may deserve some attention in future works. Some augmentation techniques lead to an unreasonably strong high-frequency bias. This frequency band contains limited information and is almost imperceivable to humans without normalization (\cref{fig:imagenet_hfc}). Nonetheless, these models seem to be able to classify a non-negligible amount of samples. This may be related to frequency shortcuts \cite{Wang_2023_ICCV} and, in fact, be less desirable.
On the other hand, we also see that augmentation improves low-frequency bias and overall the strongest performance there is achieved by an augmentation model (\textsc{DeepAugment + AugMix} \cite{hendrycks2021rendition}). This training category also contains the most models that significantly deviate from the otherwise prevalent low-frequency bias. Newer training recipes seem to mostly improve high-frequency biases, without significant changes to the low-frequency bias. SIN-only training reduces low-frequency bias but significantly raises performance in mid-bands and sometimes even outperforms augmentation on some specific cutoffs. Notably, all models make significant improvements on the lowest 1\% of the spectrum, with training recipes showing the least and AT the largest leaps. Gains from additional high-frequency information saturate much earlier for almost all models.
\begin{figure*}
    \centering
    \includegraphics[width=\linewidth]{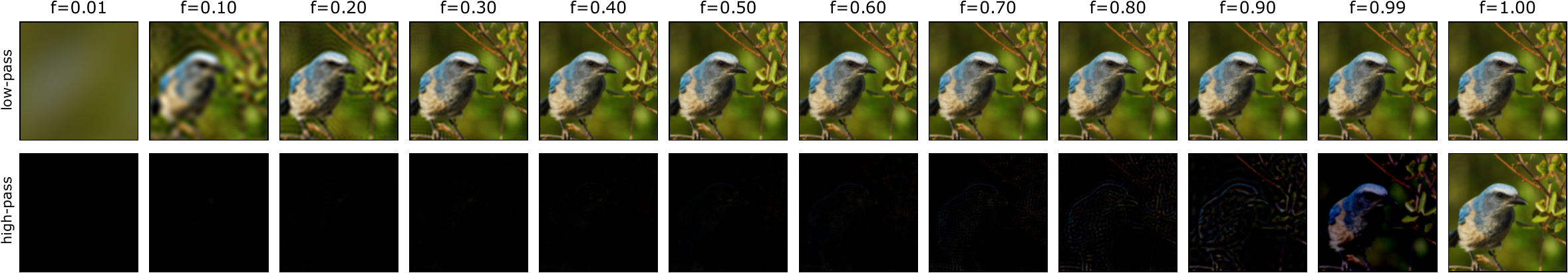}
    \caption{Visualization of the low/high-pass filtered data at cutoff frequency $f$ on one ImageNet sample.}
    \label{supp_fig:example_bandpass}
\end{figure*}
\section{Implementation Details}
\label{app_sec:implementation}
All evaluations were performed with \textit{Python} 3.10.12, \textit{PyTorch} 2.0.1, \textit{CUDA} 11.7, and \textit{cuDNN} 8500 on 4x \textit{NVIDIA A100-SXM4} GPUs.

\subsection{Data Preprocessing Pipeline}

We use the same data processing pipeline for all models to ensure a fair comparison. We resize the smaller edge of the inputs to $256$ px and the other edge with the same ratio using bilinear interpolation, then center-crop to $224\times224$ px. Channel-wise normalization is applied as done during training - typically, this is the mean and std over all samples of the ImageNet dataset. 

The samples in \cbar/C~are preprocessed by default, there we skip the the resizing and cropping. We want to point out that some prior also apply the above transformations to these datasets. As discussed in \cref{sec:method} this is questionable because it results in undersampling, and thus loss of details, and inconsistent evaluation compared to the clean ImageNet dataset and approaches using ``correct'' preprocessing.

\subsection{Implementation of the Frequency Filter}
Let $\mathcal{F}$ denote the Fast-Fourier Transformation (including shifting the zero-frequency component to the center) and $\mathcal{F}^{-1}$ the inverse operation, then we obtain the frequency filtered sample $X'$ from an input sample $X$ as follows: 
\begin{equation}
    X' = \mathcal{F}^{-1}(\mathcal{F}(X) \circ M_f)
\end{equation}
$M_c$ denotes the frequency mask (filter) in the Fourier space parameterized by the cutoff frequency $f$ implemented as follows:
\lstdefinestyle{mystyle}{
    backgroundcolor=\color{lightgray!50},   
    commentstyle=\color{codegreen},
    keywordstyle=\color{magenta},
    basicstyle=\ttfamily\footnotesize,
    breakatwhitespace=false,         
    breaklines=true,                 
    captionpos=b,                    
    keepspaces=true,                 
    numbers=left,                    
    numbersep=5pt,                  
    showspaces=false,                
    showstringspaces=false,
    showtabs=false,                  
    tabsize=2
}

\lstset{style=mystyle}

\begin{lstlisting}[language=Python, caption=Frequency Mask Computation]
h, w = X.shape[-2:]
cy, cx = h // 2, w // 2
ry = int(cutoff_freq * cy)
rx = int(cutoff_freq * cx)
if lowpass:
    mask = torch.zeros_like(X)
    mask[:, cy-ry:cy+ry, cx-rx:cx+rx] = 1
else:
    mask = torch.ones_like(X)
    mask[:, cy-ry:cy+ry, cx-rx:cx+rx] = 0
\end{lstlisting}

An example of the resulting samples can be found in \cref{supp_fig:example_bandpass}.

\begin{table*}
\centering
\caption{Performance of ResNet-50 models on our generalization benchmarks.}
\label{supp_tab:performance}
\resizebox{\linewidth}{!}{
\begin{tabular}{cllccc|ccc|ccc|c}
\toprule
&&& \multicolumn{10}{c}{\textbf{Top-1 Test Accuracy [\%]} ($\uparrow$)}\\
\cmidrule{4-13}
&&& \multicolumn{3}{c}{\textit{In Distribution}} & \multicolumn{3}{c}{\textit{Robustness}} & \multicolumn{3}{c}{\textit{Concepts}} & \multicolumn{1}{c}{\textit{Adv.}} \\
&Model &              Category &     IN &  IN-ReaL &  IN-V2 &   IN-A &   IN-C &  \cbar &   IN-R &   IN-S &    SIN &    PGD  \\
\hline 
\textcolor{c_baseline}{\scaleobj{1.0}{\bullet}}                      & Original Baseline \cite{resnet}                                                   & baseline             & 76.15 & 86.50   & 63.14 & 0.03  & 41.12 & 39.70 & 36.16 & 24.09 & 37.12 & 18.39 \\ 
\hline
\textcolor{c_adversarial_training}{\scaleobj{0.2}{\CIRCLE}}          & PGD-AT ($\ell_2$, $\epsilon$=0) \cite{salman2020transfer,madry2018towards}        & adversarial training & 75.81 & 88.65   & 63.70 & 1.85  & 40.90 & 39.48 & 35.76 & 23.50 & 34.38 & 18.88 \\
\textcolor{c_adversarial_training}{\scaleobj{0.202}{\CIRCLE}}        & PGD-AT ($\ell_2$, $\epsilon$=0.01) \cite{salman2020transfer,madry2018towards}     & adversarial training & 75.67 & 84.97   & 63.64 & 1.69  & 42.13 & 39.78 & 36.85 & 24.22 & 38.50 & 30.56 \\
\textcolor{c_adversarial_training}{\scaleobj{0.206}{\CIRCLE}}        & PGD-AT ($\ell_2$, $\epsilon$=0.03) \cite{salman2020transfer,madry2018towards}     & adversarial training & 75.77 & 87.42   & 63.33 & 1.92  & 42.25 & 39.72 & 36.71 & 24.60 & 38.50 & 40.05 \\
\textcolor{c_adversarial_training}{\scaleobj{0.21}{\CIRCLE}}         & PGD-AT ($\ell_2$, $\epsilon$=0.05) \cite{salman2020transfer,madry2018towards}     & adversarial training & 75.58 & 84.66   & 62.93 & 1.79  & 41.66 & 40.18 & 37.28 & 24.69 & 40.00 & 46.86 \\
\textcolor{c_adversarial_training}{\scaleobj{0.22}{\CIRCLE}}         & PGD-AT ($\ell_2$, $\epsilon$=0.1) \cite{salman2020transfer,madry2018towards}      & adversarial training & 74.79 & 86.20   & 62.44 & 1.88  & 41.91 & 39.35 & 37.61 & 24.70 & 39.25 & 53.76 \\
\textcolor{c_adversarial_training}{\scaleobj{0.25}{\CIRCLE}}         & PGD-AT ($\ell_2$, $\epsilon$=0.25) \cite{salman2020transfer,madry2018towards}     & adversarial training & 74.14 & 85.28   & 61.65 & 1.96  & 42.02 & 39.58 & 38.23 & 25.31 & 40.88 & 61.23 \\
\textcolor{c_adversarial_training}{\scaleobj{0.3}{\CIRCLE}}          & PGD-AT ($\ell_2$, $\epsilon$=0.5) \cite{salman2020transfer,madry2018towards}      & adversarial training & 73.17 & 86.50   & 59.97 & 2.08  & 40.82 & 39.23 & 38.94 & 24.21 & 43.88 & 64.30 \\
\textcolor{c_adversarial_training}{\scaleobj{0.4}{\CIRCLE}}          & PGD-AT ($\ell_2$, $\epsilon$=1) \cite{salman2020transfer,madry2018towards}        & adversarial training & 70.42 & 84.36   & 56.95 & 2.09  & 38.79 & 37.90 & 38.95 & 23.68 & 44.12 & 64.37 \\
\textcolor{c_adversarial_training}{\scaleobj{0.8}{\CIRCLE}}          & PGD-AT ($\ell_2$, $\epsilon$=3) \cite{salman2020transfer,madry2018towards}        & adversarial training & 62.83 & 75.77   & 48.91 & 1.87  & 34.60 & 34.83 & 36.99 & 20.93 & 41.75 & 59.47 \\
\textcolor{c_adversarial_training}{\scaleobj{1.2}{\CIRCLE}}          & PGD-AT ($\ell_2$, $\epsilon$=5) \cite{salman2020transfer,madry2018towards}        & adversarial training & 56.14 & 74.54   & 42.49 & 1.77  & 30.65 & 31.15 & 33.09 & 17.24 & 39.00 & 53.63 \\
\textcolor{c_adversarial_training}{\scaleobj{0.3}{\mdblksquare}}     & PGD-AT ($\ell_\infty$, $\epsilon$=0.5) \cite{salman2020transfer,madry2018towards} & adversarial training & 73.74 & 84.36   & 61.38 & 2.29  & 40.11 & 40.04 & 39.39 & 24.68 & 43.88 & 65.11 \\
\textcolor{c_adversarial_training}{\scaleobj{0.4}{\mdblksquare}}     & PGD-AT ($\ell_\infty$, $\epsilon$=1.0) \cite{salman2020transfer,madry2018towards} & adversarial training & 72.04 & 83.44   & 59.21 & 2.20  & 38.82 & 39.72 & 40.96 & 24.51 & 44.00 & 66.39 \\
\textcolor{c_adversarial_training}{\scaleobj{0.6}{\mdblksquare}}     & PGD-AT ($\ell_\infty$, $\epsilon$=2.0) \cite{salman2020transfer,madry2018towards} & adversarial training & 69.09 & 82.52   & 56.15 & 2.39  & 37.49 & 38.85 & 39.33 & 23.10 & 45.75 & 65.25 \\
\textcolor{c_adversarial_training}{\scaleobj{1.0}{\mdblksquare}}     & PGD-AT ($\ell_\infty$, $\epsilon$=4.0) \cite{salman2020transfer,madry2018towards} & adversarial training & 63.87 & 78.83   & 51.31 & 2.29  & 33.71 & 36.56 & 38.92 & 21.87 & 43.25 & 61.22 \\
\textcolor{c_adversarial_training}{\scaleobj{1.8}{\mdblksquare}}     & PGD-AT ($\ell_\infty$, $\epsilon$=8.0) \cite{salman2020transfer,madry2018towards} & adversarial training & 54.53 & 71.78   & 41.86 & 2.11  & 28.78 & 31.91 & 34.84 & 18.57 & 40.00 & 52.57 \\
\textcolor{c_augmentation}{\scaleobj{1.0}{\blacktriangle}}           & AugMix (180ep) \cite{hendrycks2020augmix}                                         & augmentation         & 77.53 & 88.96   & 65.42 & 3.65  & 50.77 & 46.16 & 41.03 & 28.49 & 45.50 & 30.96 \\
\textcolor{c_augmentation}{\scaleobj{1.0}{\smallblacktriangleleft}}  & DeepAugment \cite{hendrycks2021rendition}                                         & augmentation         & 76.65 & 86.81   & 65.20 & 3.40  & 54.40 & 48.39 & 42.25 & 29.50 & 49.12 & 32.51 \\
\textcolor{c_augmentation}{\scaleobj{1.0}{\smallblacktriangleright}} & DeepAugment+AugMix \cite{hendrycks2021rendition}                                  & augmentation         & 75.80 & 86.20   & 63.65 & 3.85  & 59.53 & 51.34 & 46.79 & 32.62 & 57.50 & 40.40 \\
\textcolor{c_augmentation}{\scaleobj{1.0}{\bullet}}                  & Noise Training (clean eval) \cite{jaini2023intriguing}                            & augmentation         & 67.22 & 83.44   & 54.67 & 2.43  & 44.40 & 39.48 & 36.64 & 19.99 & 47.12 & 48.27 \\
\textcolor{c_augmentation}{\scaleobj{1.0}{\btimes}}                  & NoisyMix \cite{erichson2022noisymix}                                              & augmentation         & 77.05 & 89.57   & 64.28 & 3.32  & 54.23 & 50.62 & 45.77 & 31.18 & 49.38 & 50.70 \\
\textcolor{c_augmentation}{\scaleobj{1.0}{\CIRCLE}}                  & OpticsAugment \cite{Muller_2023_ICCV}                                             & augmentation         & 74.22 & 86.50   & 62.03 & 1.73  & 42.90 & 40.39 & 37.50 & 24.69 & 43.88 & 16.08 \\
\textcolor{c_augmentation}{\scaleobj{1.0}{\blacklozenge}}            & PRIME \cite{modos2022prime}                                                       & augmentation         & 76.91 & 87.12   & 64.34 & 2.16  & 55.27 & 49.00 & 42.20 & 29.83 & 46.62 & 30.82 \\
\textcolor{c_augmentation}{\scaleobj{1.0}{\blackoctagon}}            & PixMix (180ep) \cite{hendrycks2022pixmix}                                         & augmentation         & 78.09 & 88.65   & 65.89 & 6.25  & 52.99 & 59.51 & 40.31 & 29.21 & 40.25 & 23.02 \\
\textcolor{c_augmentation}{\scaleobj{1.0}{\pentagofill}}             & PixMix (90ep) \cite{hendrycks2022pixmix}                                          & augmentation         & 77.36 & 89.88   & 65.20 & 4.11  & 51.87 & 57.76 & 39.92 & 28.57 & 45.00 & 22.28 \\
\textcolor{c_augmentation}{\scaleobj{1.0}{\varhexagonblack}}         & Shape Bias Augmentation \cite{li2021shapetexture}                                 & augmentation         & 76.21 & 87.42   & 64.20 & 3.03  & 47.60 & 44.46 & 40.64 & 27.92 & 64.50 & 25.18 \\
\textcolor{c_augmentation}{\scaleobj{1.0}{\triup}}                   & Texture Bias Augmentation \cite{li2021shapetexture}                               & augmentation         & 75.27 & 86.81   & 63.18 & 2.25  & 41.82 & 40.26 & 36.76 & 24.28 & 35.50 & 16.83 \\
\textcolor{c_augmentation}{\scaleobj{1.0}{\tridown}}                 & Texture/Shape Debiased Augmentation \cite{li2021shapetexture}                     & augmentation         & 76.89 & 86.20   & 65.04 & 3.39  & 48.28 & 45.47 & 40.77 & 28.42 & 56.00 & 25.99 \\
\textcolor{c_contrastive}{\scaleobj{1.0}{\varhexagonblack}}          & DINO V1 \cite{caron2021emerging}                                                  & contrastive          & 75.28 & 85.28   & 62.70 & 5.15  & 39.61 & 35.88 & 30.17 & 18.75 & 30.63 & 13.26 \\
\textcolor{c_contrastive}{\scaleobj{1.0}{\btimes}}                   & MoCo V3 (1000ep) \cite{Chen_2021_ICCV}                                            & contrastive          & 74.60 & 87.42   & 62.01 & 4.07  & 43.53 & 40.76 & 37.05 & 25.51 & 35.50 & 27.79 \\
\textcolor{c_contrastive}{\scaleobj{1.0}{\tridown}}                  & MoCo V3 (100ep) \cite{Chen_2021_ICCV}                                             & contrastive          & 68.91 & 82.52   & 56.28 & 2.43  & 37.75 & 36.62 & 31.71 & 20.48 & 36.75 & 24.61 \\
\textcolor{c_contrastive}{\scaleobj{1.0}{\triup}}                    & MoCo V3 (300ep) \cite{Chen_2021_ICCV}                                             & contrastive          & 72.80 & 84.97   & 60.74 & 3.27  & 41.97 & 39.00 & 35.41 & 24.00 & 36.75 & 27.57 \\
\textcolor{c_contrastive}{\scaleobj{1.0}{\bullet}}                   & SimCLRv2 \cite{chen2020simclrv2}                                                  & contrastive          & 74.90 & 85.58   & 61.24 & 4.65  & 44.32 & 40.73 & 35.16 & 23.55 & 43.88 & 14.57 \\
\textcolor{c_contrastive}{\scaleobj{1.0}{\CIRCLE}}                   & SwAV \cite{caron2020swav}                                                         & contrastive          & 75.31 & 87.73   & 62.15 & 5.49  & 41.48 & 37.63 & 30.24 & 18.94 & 30.38 & 14.75 \\
\textcolor{c_freezing}{\scaleobj{1.0}{\bullet}}                      & Frozen Random Filters \cite{gavrikov2023power}                                    & freezing             & 74.76 & 87.12   & 62.47 & 2.52  & 45.22 & 40.98 & 37.52 & 25.36 & 40.62 & 16.18 \\
\textcolor{c_stylized}{\scaleobj{1.0}{\varhexagonblack}}             & ShapeNet: SIN Training \cite{geirhos2018imagenettrained}                          & stylized             & 60.18 & 73.31   & 48.61 & 2.39  & 39.76 & 36.76 & 40.17 & 30.09 & 90.88 & 12.33 \\
\textcolor{c_stylized}{\scaleobj{1.0}{\bullet}}                      & ShapeNet: SIN+IN Training \cite{geirhos2018imagenettrained}                       & stylized             & 74.59 & 87.12   & 62.43 & 1.91  & 46.91 & 43.33 & 41.55 & 29.70 & 91.00 & 22.47 \\
\textcolor{c_stylized}{\scaleobj{1.0}{\CIRCLE}}                      & ShapeNet: SIN+IN Training + FT \cite{geirhos2018imagenettrained}                  & stylized             & 76.72 & 88.34   & 64.65 & 2.23  & 43.55 & 41.86 & 38.93 & 26.92 & 45.00 & 21.23 \\
\textcolor{c_training_recipes}{\scaleobj{1.0}{\blackoctagon}}        & timm (A1) \cite{rw2019timm,wightman2021resnet}                                    & training recipes     & 80.10 & 88.65   & 68.73 & 11.03 & 50.93 & 49.01 & 40.60 & 29.22 & 36.88 & 27.74 \\
\textcolor{c_training_recipes}{\scaleobj{1.0}{\CIRCLE}}              & timm (A1H) \cite{rw2019timm,wightman2021resnet}                                   & training recipes     & 80.10 & 89.26   & 68.47 & 15.21 & 49.36 & 48.57 & 40.99 & 29.64 & 39.00 & 36.11 \\
\textcolor{c_training_recipes}{\scaleobj{1.0}{\pentagofill}}         & timm (A2) \cite{rw2019timm,wightman2021resnet}                                    & training recipes     & 79.80 & 86.20   & 67.29 & 7.36  & 48.98 & 47.83 & 38.39 & 27.27 & 38.38 & 27.03 \\
\textcolor{c_training_recipes}{\scaleobj{1.0}{\bullet}}              & timm (A3) \cite{rw2019timm,wightman2021resnet}                                    & training recipes     & 77.55 & 86.81   & 65.04 & 6.35  & 41.03 & 43.20 & 35.93 & 24.61 & 34.75 & 23.32 \\
\textcolor{c_training_recipes}{\scaleobj{1.0}{\btimes}}              & timm (B1K) \cite{rw2019timm,wightman2021resnet}                                   & training recipes     & 79.16 & 88.34   & 67.41 & 8.51  & 51.64 & 50.30 & 43.04 & 31.22 & 42.88 & 33.40 \\
\textcolor{c_training_recipes}{\scaleobj{1.0}{\triup}}               & timm (B2K) \cite{rw2019timm,wightman2021resnet}                                   & training recipes     & 79.27 & 87.42   & 67.79 & 8.64  & 52.25 & 50.05 & 42.44 & 30.40 & 40.75 & 32.82 \\
\textcolor{c_training_recipes}{\scaleobj{1.0}{\blacklozenge}}        & timm (C1) \cite{rw2019timm,wightman2021resnet}                                    & training recipes     & 79.76 & 89.88   & 68.54 & 10.07 & 50.60 & 49.40 & 41.54 & 30.29 & 37.12 & 33.72 \\
\textcolor{c_training_recipes}{\scaleobj{1.0}{\tridown}}             & timm (C2) \cite{rw2019timm,wightman2021resnet}                                    & training recipes     & 79.92 & 90.49   & 68.80 & 11.49 & 51.62 & 50.92 & 40.73 & 29.85 & 37.12 & 30.38 \\
\textcolor{c_training_recipes}{\scaleobj{1.0}{\varhexagonblack}}     & timm (D) \cite{rw2019timm,wightman2021resnet}                                     & training recipes     & 79.89 & 89.26   & 68.73 & 9.76  & 51.26 & 49.53 & 40.61 & 29.85 & 36.00 & 29.62 \\
\textcolor{c_training_recipes}{\scaleobj{1.0}{\blacktriangle}}       & torchvision (V2) \cite{BibEntry2023Nov,PyTorch}                                   & training recipes     & 80.34 & 90.18   & 69.57 & 16.73 & 50.02 & 49.67 & 41.62 & 28.44 & 38.38 & 39.90\\
\bottomrule
\end{tabular}
}
\end{table*}

\begin{table*}
\centering
\caption{Bias measurements of ResNet-50 models. Columns with gray background were not directly used in the main paper.}
\label{sup_tab:influences}
\resizebox{\linewidth}{!}{
\begin{tabular}{cll|r|rr|>{\columncolor[gray]{0.8}}r>{\columncolor[gray]{0.8}}r>{\columncolor[gray]{0.8}}r|rrr}
\toprule
                                         &                                                                                   &                      & \multicolumn{9}{c}{\textbf{Bias}}                                                                                                                                                                                                              \\ 
\cmidrule{4-12}
\multicolumn{1}{l}{}                     &                                                                                   &                      & \multicolumn{1}{l|}{} & \multicolumn{1}{l}{} & \multicolumn{1}{l|}{} & \multicolumn{6}{c}{Critical Band}                                                                                                                                       \\
                                         &                                                                                   &                      & Shape                 & \multicolumn{2}{c|}{Spectral}                & \multicolumn{3}{c|}{\textit{(IN-1k non-normalized)}}                                                                 & \multicolumn{3}{c}{\textit{(IN-1k normalized)}}  \\
                                         & Model                                                                             & Category             & \multicolumn{1}{c|}{Bias}                   & Low-Freq.            & High-Freq.            & {\cellcolor[rgb]{0.8,0.8,0.8}}C-BW     & {\cellcolor[rgb]{0.8,0.8,0.8}}C-CF    & {\cellcolor[rgb]{0.8,0.8,0.8}}C-PNS & C-BW  & C-CF   & C-PNS                           \\ 
\midrule
\scaleobj{1.0}{\textcolor{c_baseline}\bullet}                  & Original Baseline \cite{resnet}                                                   & baseline             & 0.21                  & 0.63                 & 0.01                  & {\cellcolor[rgb]{0.8,0.8,0.8}}11295.35 & {\cellcolor[rgb]{0.8,0.8,0.8}}3681.95 & {\cellcolor[rgb]{0.8,0.8,0.8}}1.0   & 5.67  & 54.68  & 1.00                            \\ 
\hline
\scaleobj{0.2}{\textcolor{c_adversarial_training}\CIRCLE}                  & PGD-AT ($\ell_2$, $\epsilon$=0) \cite{salman2020transfer,madry2018towards}        & adversarial training & 0.21                  & 0.62                 & 0.01                  & {\cellcolor[rgb]{0.8,0.8,0.8}}11295.35 & {\cellcolor[rgb]{0.8,0.8,0.8}}3681.95 & {\cellcolor[rgb]{0.8,0.8,0.8}}1.0   & 5.67  & 54.68  & 1.00                            \\
\scaleobj{0.202}{\textcolor{c_adversarial_training}\CIRCLE}                & PGD-AT ($\ell_2$, $\epsilon$=0.01) \cite{salman2020transfer,madry2018towards}     & adversarial training & 0.22                  & 0.64                 & 0.01                  & {\cellcolor[rgb]{0.8,0.8,0.8}}11295.35 & {\cellcolor[rgb]{0.8,0.8,0.8}}3681.95 & {\cellcolor[rgb]{0.8,0.8,0.8}}1.0   & 5.67  & 54.68  & 1.00                            \\
\scaleobj{0.206}{\textcolor{c_adversarial_training}\CIRCLE}                & PGD-AT ($\ell_2$, $\epsilon$=0.03) \cite{salman2020transfer,madry2018towards}     & adversarial training & 0.24                  & 0.67                 & 0.00                  & {\cellcolor[rgb]{0.8,0.8,0.8}}11295.35 & {\cellcolor[rgb]{0.8,0.8,0.8}}3681.95 & {\cellcolor[rgb]{0.8,0.8,0.8}}1.0   & 5.67  & 54.68  & 1.00                            \\
\scaleobj{0.21}{\textcolor{c_adversarial_training}\CIRCLE}                 & PGD-AT ($\ell_2$, $\epsilon$=0.05) \cite{salman2020transfer,madry2018towards}     & adversarial training & 0.24                  & 0.67                 & 0.00                  & {\cellcolor[rgb]{0.8,0.8,0.8}}11295.35 & {\cellcolor[rgb]{0.8,0.8,0.8}}3681.95 & {\cellcolor[rgb]{0.8,0.8,0.8}}1.0   & 5.91  & 41.85  & 1.00                            \\
\scaleobj{0.22}{\textcolor{c_adversarial_training}\CIRCLE}                 & PGD-AT ($\ell_2$, $\epsilon$=0.1) \cite{salman2020transfer,madry2018towards}      & adversarial training & 0.28                  & 0.69                 & 0.00                  & {\cellcolor[rgb]{0.8,0.8,0.8}}11295.35 & {\cellcolor[rgb]{0.8,0.8,0.8}}3681.95 & {\cellcolor[rgb]{0.8,0.8,0.8}}1.0   & 6.45  & 40.65  & 1.00                            \\
\scaleobj{0.25}{\textcolor{c_adversarial_training}\CIRCLE}                 & PGD-AT ($\ell_2$, $\epsilon$=0.25) \cite{salman2020transfer,madry2018towards}     & adversarial training & 0.34                  & 0.72                 & 0.00                  & {\cellcolor[rgb]{0.8,0.8,0.8}}11295.35 & {\cellcolor[rgb]{0.8,0.8,0.8}}3681.95 & {\cellcolor[rgb]{0.8,0.8,0.8}}1.0   & 7.86  & 58.20  & 1.00                            \\
\scaleobj{0.3}{\textcolor{c_adversarial_training}\CIRCLE}                  & PGD-AT ($\ell_2$, $\epsilon$=0.5) \cite{salman2020transfer,madry2018towards}      & adversarial training & 0.41                  & 0.73                 & 0.00                  & {\cellcolor[rgb]{0.8,0.8,0.8}}11295.35 & {\cellcolor[rgb]{0.8,0.8,0.8}}3681.95 & {\cellcolor[rgb]{0.8,0.8,0.8}}1.0   & 7.86  & 58.20  & 1.00                            \\
\scaleobj{0.4}{\textcolor{c_adversarial_training}\CIRCLE}                  & PGD-AT ($\ell_2$, $\epsilon$=1) \cite{salman2020transfer,madry2018towards}        & adversarial training & 0.48                  & 0.75                 & 0.00                  & {\cellcolor[rgb]{0.8,0.8,0.8}}11295.35 & {\cellcolor[rgb]{0.8,0.8,0.8}}3681.95 & {\cellcolor[rgb]{0.8,0.8,0.8}}1.0   & 7.86  & 58.20  & 1.00                            \\
\scaleobj{0.8}{\textcolor{c_adversarial_training}\CIRCLE}                  & PGD-AT ($\ell_2$, $\epsilon$=3) \cite{salman2020transfer,madry2018towards}        & adversarial training & 0.65                  & 0.76                 & 0.00                  & {\cellcolor[rgb]{0.8,0.8,0.8}}11295.35 & {\cellcolor[rgb]{0.8,0.8,0.8}}3681.95 & {\cellcolor[rgb]{0.8,0.8,0.8}}1.0   & 7.47  & 32.63  & 0.60                            \\
\scaleobj{1.2}{\textcolor{c_adversarial_training}\CIRCLE}                  & PGD-AT ($\ell_2$, $\epsilon$=5) \cite{salman2020transfer,madry2018towards}        & adversarial training & 0.69                  & 0.78                 & 0.00                  & {\cellcolor[rgb]{0.8,0.8,0.8}}11295.35 & {\cellcolor[rgb]{0.8,0.8,0.8}}3681.95 & {\cellcolor[rgb]{0.8,0.8,0.8}}1.0   & 7.92  & 45.71  & 0.55                            \\
\scaleobj{0.3}{\textcolor{c_adversarial_training}\mdblksquare}             & PGD-AT ($\ell_\infty$, $\epsilon$=0.5) \cite{salman2020transfer,madry2018towards} & adversarial training & 0.37                  & 0.73                 & 0.00                  & {\cellcolor[rgb]{0.8,0.8,0.8}}11295.35 & {\cellcolor[rgb]{0.8,0.8,0.8}}3681.95 & {\cellcolor[rgb]{0.8,0.8,0.8}}1.0   & 7.86  & 58.20  & 1.00                            \\
\scaleobj{0.4}{\textcolor{c_adversarial_training}\mdblksquare}             & PGD-AT ($\ell_\infty$, $\epsilon$=1.0) \cite{salman2020transfer,madry2018towards} & adversarial training & 0.45                  & 0.74                 & 0.00                  & {\cellcolor[rgb]{0.8,0.8,0.8}}11295.35 & {\cellcolor[rgb]{0.8,0.8,0.8}}3681.95 & {\cellcolor[rgb]{0.8,0.8,0.8}}1.0   & 7.37  & 39.19  & 0.73                            \\
\scaleobj{0.6}{\textcolor{c_adversarial_training}\mdblksquare}             & PGD-AT ($\ell_\infty$, $\epsilon$=2.0) \cite{salman2020transfer,madry2018towards} & adversarial training & 0.54                  & 0.75                 & 0.00                  & {\cellcolor[rgb]{0.8,0.8,0.8}}11295.35 & {\cellcolor[rgb]{0.8,0.8,0.8}}3681.95 & {\cellcolor[rgb]{0.8,0.8,0.8}}1.0   & 7.92  & 45.71  & 0.55                            \\
\scaleobj{1.0}{\textcolor{c_adversarial_training}\mdblksquare}             & PGD-AT ($\ell_\infty$, $\epsilon$=4.0) \cite{salman2020transfer,madry2018towards} & adversarial training & 0.62                  & 0.75                 & 0.00                  & {\cellcolor[rgb]{0.8,0.8,0.8}}11295.35 & {\cellcolor[rgb]{0.8,0.8,0.8}}3681.95 & {\cellcolor[rgb]{0.8,0.8,0.8}}1.0   & 10.46 & 114.50 & 0.52                            \\
\scaleobj{1.8}{\textcolor{c_adversarial_training}\mdblksquare}             & PGD-AT ($\ell_\infty$, $\epsilon$=8.0) \cite{salman2020transfer,madry2018towards} & adversarial training & 0.72                  & 0.78                 & 0.00                  & {\cellcolor[rgb]{0.8,0.8,0.8}}11295.35 & {\cellcolor[rgb]{0.8,0.8,0.8}}3681.95 & {\cellcolor[rgb]{0.8,0.8,0.8}}1.0   & 7.47  & 32.63  & 0.60                            \\
\scaleobj{1.0}{\textcolor{c_augmentation}\blacktriangle}           & AugMix (180ep) \cite{hendrycks2020augmix}                                         & augmentation         & 0.30                  & 0.74                 & 0.02                  & {\cellcolor[rgb]{0.8,0.8,0.8}}9.95     & {\cellcolor[rgb]{0.8,0.8,0.8}}34.70   & {\cellcolor[rgb]{0.8,0.8,0.8}}1.0   & 4.74  & 38.85  & 1.00                            \\
\scaleobj{1.0}{\textcolor{c_augmentation}\smallblacktriangleleft}  & DeepAugment \cite{hendrycks2021rendition}                                         & augmentation         & 0.39                  & 0.77                 & 0.06                  & {\cellcolor[rgb]{0.8,0.8,0.8}}11295.35 & {\cellcolor[rgb]{0.8,0.8,0.8}}3681.95 & {\cellcolor[rgb]{0.8,0.8,0.8}}1.0   & 5.51  & 56.31  & 0.72                            \\
\scaleobj{1.0}{\textcolor{c_augmentation}\smallblacktriangleright} & DeepAugment+AugMix \cite{hendrycks2021rendition}                                  & augmentation         & 0.52                  & 0.84                 & 0.09                  & {\cellcolor[rgb]{0.8,0.8,0.8}}9.95     & {\cellcolor[rgb]{0.8,0.8,0.8}}34.70   & {\cellcolor[rgb]{0.8,0.8,0.8}}1.0   & 4.67  & 38.28  & 0.63                            \\
\scaleobj{1.0}{\textcolor{c_augmentation}\bullet}                  & Noise Training (clean eval) \cite{jaini2023intriguing}                            & augmentation         & 0.51                  & 0.79                 & 0.01                  & {\cellcolor[rgb]{0.8,0.8,0.8}}11295.35 & {\cellcolor[rgb]{0.8,0.8,0.8}}3681.95 & {\cellcolor[rgb]{0.8,0.8,0.8}}1.0   & 6.35  & 33.58  & 0.93                            \\
\scaleobj{1.0}{\textcolor{c_augmentation}\btimes}                  & NoisyMix \cite{erichson2022noisymix}                                              & augmentation         & 0.32                  & 0.75                 & 0.01                  & {\cellcolor[rgb]{0.8,0.8,0.8}}9.95     & {\cellcolor[rgb]{0.8,0.8,0.8}}34.70   & {\cellcolor[rgb]{0.8,0.8,0.8}}1.0   & 4.74  & 38.85  & 1.00                            \\
\scaleobj{1.0}{\textcolor{c_augmentation}\CIRCLE}                  & OpticsAugment \cite{Muller_2023_ICCV}                                             & augmentation         & 0.24                  & 0.63                 & 0.01                  & {\cellcolor[rgb]{0.8,0.8,0.8}}11295.35 & {\cellcolor[rgb]{0.8,0.8,0.8}}3681.95 & {\cellcolor[rgb]{0.8,0.8,0.8}}1.0   & 4.45  & 61.21  & 1.00                            \\
\scaleobj{1.0}{\textcolor{c_augmentation}\blacklozenge}            & PRIME \cite{modos2022prime}                                                       & augmentation         & 0.32                  & 0.71                 & 0.13                  & {\cellcolor[rgb]{0.8,0.8,0.8}}7.34     & {\cellcolor[rgb]{0.8,0.8,0.8}}42.82   & {\cellcolor[rgb]{0.8,0.8,0.8}}1.0   & 4.74  & 38.85  & 1.00                            \\
\scaleobj{1.0}{\textcolor{c_augmentation}\blackoctagon}            & PixMix (180ep) \cite{hendrycks2022pixmix}                                         & augmentation         & 0.26                  & 0.68                 & 0.03                  & {\cellcolor[rgb]{0.8,0.8,0.8}}5.91     & {\cellcolor[rgb]{0.8,0.8,0.8}}41.85   & {\cellcolor[rgb]{0.8,0.8,0.8}}1.0   & 4.74  & 38.85  & 1.00                            \\
\scaleobj{1.0}{\textcolor{c_augmentation}\pentagofill}             & PixMix (90ep) \cite{hendrycks2022pixmix}                                          & augmentation         & 0.23                  & 0.67                 & 0.03                  & {\cellcolor[rgb]{0.8,0.8,0.8}}6.45     & {\cellcolor[rgb]{0.8,0.8,0.8}}40.65   & {\cellcolor[rgb]{0.8,0.8,0.8}}1.0   & 4.74  & 38.85  & 1.00                            \\
\scaleobj{1.0}{\textcolor{c_augmentation}\varhexagonblack}         & Shape Bias Augmentation \cite{li2021shapetexture}                                 & augmentation         & 0.28                  & 0.66                 & 0.01                  & {\cellcolor[rgb]{0.8,0.8,0.8}}11295.35 & {\cellcolor[rgb]{0.8,0.8,0.8}}3681.95 & {\cellcolor[rgb]{0.8,0.8,0.8}}1.0   & 4.74  & 38.85  & 1.00                            \\
\scaleobj{1.0}{\textcolor{c_augmentation}\triup}                   & Texture Bias Augmentation \cite{li2021shapetexture}                               & augmentation         & 0.20                  & 0.64                 & 0.01                  & {\cellcolor[rgb]{0.8,0.8,0.8}}11295.35 & {\cellcolor[rgb]{0.8,0.8,0.8}}3681.95 & {\cellcolor[rgb]{0.8,0.8,0.8}}1.0   & 5.67  & 54.68  & 1.00                            \\
\scaleobj{1.0}{\textcolor{c_augmentation}\tridown}                 & Texture/Shape Debiased Augmentation \cite{li2021shapetexture}                     & augmentation         & 0.26                  & 0.67                 & 0.01                  & {\cellcolor[rgb]{0.8,0.8,0.8}}11295.35 & {\cellcolor[rgb]{0.8,0.8,0.8}}3681.95 & {\cellcolor[rgb]{0.8,0.8,0.8}}1.0   & 5.67  & 54.68  & 1.00                            \\
\scaleobj{1.0}{\textcolor{c_contrastive}\varhexagonblack}         & DINO V1 \cite{caron2021emerging}                                                  & contrastive          & 0.18                  & 0.44                 & 0.01                  & {\cellcolor[rgb]{0.8,0.8,0.8}}7.34     & {\cellcolor[rgb]{0.8,0.8,0.8}}42.82   & {\cellcolor[rgb]{0.8,0.8,0.8}}1.0   & 4.74  & 38.85  & 1.00                            \\
\scaleobj{1.0}{\textcolor{c_contrastive}\btimes}                  & MoCo V3 (1000ep) \cite{Chen_2021_ICCV}                                            & contrastive          & 0.33                  & 0.56                 & 0.01                  & {\cellcolor[rgb]{0.8,0.8,0.8}}7.34     & {\cellcolor[rgb]{0.8,0.8,0.8}}42.82   & {\cellcolor[rgb]{0.8,0.8,0.8}}1.0   & 4.74  & 38.85  & 1.00                            \\
\scaleobj{1.0}{\textcolor{c_contrastive}\tridown}                 & MoCo V3 (100ep) \cite{Chen_2021_ICCV}                                             & contrastive          & 0.30                  & 0.55                 & 0.01                  & {\cellcolor[rgb]{0.8,0.8,0.8}}8.22     & {\cellcolor[rgb]{0.8,0.8,0.8}}33.66   & {\cellcolor[rgb]{0.8,0.8,0.8}}1.0   & 4.74  & 38.85  & 1.00                            \\
\scaleobj{1.0}{\textcolor{c_contrastive}\triup}                   & MoCo V3 (300ep) \cite{Chen_2021_ICCV}                                             & contrastive          & 0.31                  & 0.55                 & 0.01                  & {\cellcolor[rgb]{0.8,0.8,0.8}}7.34     & {\cellcolor[rgb]{0.8,0.8,0.8}}42.82   & {\cellcolor[rgb]{0.8,0.8,0.8}}1.0   & 4.74  & 38.85  & 1.00                            \\
\scaleobj{1.0}{\textcolor{c_contrastive}\bullet}                  & SimCLRv2 \cite{chen2020simclrv2}                                                  & contrastive          & 0.23                  & 0.55                 & 0.01                  & {\cellcolor[rgb]{0.8,0.8,0.8}}11295.35 & {\cellcolor[rgb]{0.8,0.8,0.8}}3681.95 & {\cellcolor[rgb]{0.8,0.8,0.8}}1.0   & 4.74  & 38.85  & 1.00                            \\
\scaleobj{1.0}{\textcolor{c_contrastive}\CIRCLE}                  & SwAV \cite{caron2020swav}                                                         & contrastive          & 0.18                  & 0.43                 & 0.01                  & {\cellcolor[rgb]{0.8,0.8,0.8}}7.34     & {\cellcolor[rgb]{0.8,0.8,0.8}}42.82   & {\cellcolor[rgb]{0.8,0.8,0.8}}1.0   & 4.74  & 38.85  & 1.00                            \\
\scaleobj{1.0}{\textcolor{c_freezing}\bullet}                  & Frozen Random Filters \cite{gavrikov2023power}                                    & freezing             & 0.31                  & 0.68                 & 0.01                  & {\cellcolor[rgb]{0.8,0.8,0.8}}11295.35 & {\cellcolor[rgb]{0.8,0.8,0.8}}3681.95 & {\cellcolor[rgb]{0.8,0.8,0.8}}1.0   & 4.74  & 38.85  & 1.00                            \\
\scaleobj{1.0}{\textcolor{c_stylized}\varhexagonblack}         & ShapeNet: SIN Training \cite{geirhos2018imagenettrained}                          & stylized             & 0.81                  & 0.56                 & 0.04                  & {\cellcolor[rgb]{0.8,0.8,0.8}}11295.35 & {\cellcolor[rgb]{0.8,0.8,0.8}}3681.95 & {\cellcolor[rgb]{0.8,0.8,0.8}}1.0   & 4.42  & 45.57  & 0.67                            \\
\scaleobj{1.0}{\textcolor{c_stylized}\bullet}                  & ShapeNet: SIN+IN Training \cite{geirhos2018imagenettrained}                       & stylized             & 0.35                  & 0.63                 & 0.01                  & {\cellcolor[rgb]{0.8,0.8,0.8}}11295.35 & {\cellcolor[rgb]{0.8,0.8,0.8}}3681.95 & {\cellcolor[rgb]{0.8,0.8,0.8}}1.0   & 4.33  & 39.67  & 1.00                            \\
\scaleobj{1.0}{\textcolor{c_stylized}\CIRCLE}                  & ShapeNet: SIN+IN Training + FT \cite{geirhos2018imagenettrained}                  & stylized             & 0.20                  & 0.64                 & 0.01                  & {\cellcolor[rgb]{0.8,0.8,0.8}}11295.35 & {\cellcolor[rgb]{0.8,0.8,0.8}}3681.95 & {\cellcolor[rgb]{0.8,0.8,0.8}}1.0   & 5.67  & 54.68  & 1.00                            \\
\scaleobj{1.0}{\textcolor{c_training_recipes}\blackoctagon}            & timm (A1) \cite{rw2019timm,wightman2021resnet}                                    & training recipes     & 0.21                  & 0.63                 & 0.02                  & {\cellcolor[rgb]{0.8,0.8,0.8}}5.67     & {\cellcolor[rgb]{0.8,0.8,0.8}}54.68   & {\cellcolor[rgb]{0.8,0.8,0.8}}1.0   & 3.96  & 42.09  & 1.00                            \\
\scaleobj{1.0}{\textcolor{c_training_recipes}\CIRCLE}                  & timm (A1H) \cite{rw2019timm,wightman2021resnet}                                   & training recipes     & 0.17                  & 0.61                 & 0.02                  & {\cellcolor[rgb]{0.8,0.8,0.8}}5.67     & {\cellcolor[rgb]{0.8,0.8,0.8}}54.68   & {\cellcolor[rgb]{0.8,0.8,0.8}}1.0   & 3.71  & 46.56  & 1.00                            \\
\scaleobj{1.0}{\textcolor{c_training_recipes}\pentagofill}             & timm (A2) \cite{rw2019timm,wightman2021resnet}                                    & training recipes     & 0.16                  & 0.62                 & 0.01                  & {\cellcolor[rgb]{0.8,0.8,0.8}}5.67     & {\cellcolor[rgb]{0.8,0.8,0.8}}54.68   & {\cellcolor[rgb]{0.8,0.8,0.8}}1.0   & 4.33  & 39.67  & 1.00                            \\
\scaleobj{1.0}{\textcolor{c_training_recipes}\bullet}                  & timm (A3) \cite{rw2019timm,wightman2021resnet}                                    & training recipes     & 0.13                  & 0.58                 & 0.01                  & {\cellcolor[rgb]{0.8,0.8,0.8}}9.95     & {\cellcolor[rgb]{0.8,0.8,0.8}}34.70   & {\cellcolor[rgb]{0.8,0.8,0.8}}1.0   & 4.67  & 54.89  & 1.00                            \\
\scaleobj{1.0}{\textcolor{c_training_recipes}\btimes}                  & timm (B1K) \cite{rw2019timm,wightman2021resnet}                                   & training recipes     & 0.19                  & 0.64                 & 0.02                  & {\cellcolor[rgb]{0.8,0.8,0.8}}5.67     & {\cellcolor[rgb]{0.8,0.8,0.8}}54.68   & {\cellcolor[rgb]{0.8,0.8,0.8}}1.0   & 4.14  & 47.02  & 0.91                            \\
\scaleobj{1.0}{\textcolor{c_training_recipes}\triup}                   & timm (B2K) \cite{rw2019timm,wightman2021resnet}                                   & training recipes     & 0.18                  & 0.65                 & 0.02                  & {\cellcolor[rgb]{0.8,0.8,0.8}}5.67     & {\cellcolor[rgb]{0.8,0.8,0.8}}54.68   & {\cellcolor[rgb]{0.8,0.8,0.8}}1.0   & 4.14  & 47.02  & 0.91                            \\
\scaleobj{1.0}{\textcolor{c_training_recipes}\blacklozenge}            & timm (C1) \cite{rw2019timm,wightman2021resnet}                                    & training recipes     & 0.18                  & 0.64                 & 0.02                  & {\cellcolor[rgb]{0.8,0.8,0.8}}5.67     & {\cellcolor[rgb]{0.8,0.8,0.8}}54.68   & {\cellcolor[rgb]{0.8,0.8,0.8}}1.0   & 3.96  & 42.09  & 1.00                            \\
\scaleobj{1.0}{\textcolor{c_training_recipes}\tridown}                 & timm (C2) \cite{rw2019timm,wightman2021resnet}                                    & training recipes     & 0.18                  & 0.62                 & 0.02                  & {\cellcolor[rgb]{0.8,0.8,0.8}}5.67     & {\cellcolor[rgb]{0.8,0.8,0.8}}54.68   & {\cellcolor[rgb]{0.8,0.8,0.8}}1.0   & 3.92  & 53.43  & 0.86                            \\
\scaleobj{1.0}{\textcolor{c_training_recipes}\varhexagonblack}         & timm (D) \cite{rw2019timm,wightman2021resnet}                                     & training recipes     & 0.17                  & 0.63                 & 0.01                  & {\cellcolor[rgb]{0.8,0.8,0.8}}5.67     & {\cellcolor[rgb]{0.8,0.8,0.8}}54.68   & {\cellcolor[rgb]{0.8,0.8,0.8}}1.0   & 3.96  & 42.09  & 1.00                            \\
\scaleobj{1.0}{\textcolor{c_training_recipes}\blacktriangle}           & torchvision (V2) \cite{BibEntry2023Nov,PyTorch}                                   & training recipes     & 0.17                  & 0.66                 & 0.01                  & {\cellcolor[rgb]{0.8,0.8,0.8}}5.67     & {\cellcolor[rgb]{0.8,0.8,0.8}}54.68   & {\cellcolor[rgb]{0.8,0.8,0.8}}1.0   & 3.92  & 53.43  & 0.86                            \\
\bottomrule
\end{tabular}
}
\end{table*}

\end{document}